\documentclass[lettersize,journal]{IEEEtran}
\usepackage{amsmath,amsfonts}
\usepackage{algorithmic}
\usepackage{algorithm}
\usepackage{array}
\usepackage[caption=false,font=normalsize,labelfont=sf,textfont=sf]{subfig}
\usepackage{textcomp}
\usepackage{stfloats}
\usepackage{url}
\usepackage{verbatim}
\usepackage{graphicx}
\usepackage{cite}
\usepackage{multirow}
\usepackage{makecell}
\usepackage{color}
\usepackage{soul}
\usepackage{ulem}
\usepackage{color}
\usepackage{colortbl}
\usepackage{hyperref}
\begin{document}
	\sloppy
	\definecolor{black}{rgb}{0.0, 0.0, 0.0}
	\definecolor{white}{rgb}{1.0, 1.0, 1.0}
	\definecolor{yellow}{rgb}{1.0, 1.0, 0.8}
	\definecolor{green}{rgb}{1, 0.9, 0.8}
	\definecolor{dark_green}{RGB} {0, 140, 0}
	\definecolor{gold}{rgb}{0.6, 0.4, 0.1}
	\definecolor{grey}{RGB}{0,0,0}
	\definecolor{Gray}{gray}{0.8}
	\definecolor{MediumGray}{gray}{0.9}
	\definecolor{LightGray}{gray}{0.98}
	\definecolor{LightCyan}{rgb}{0.88,1,1}
	\definecolor{purple}{RGB}{128,0,128}
	\definecolor{sl_blue}{RGB}{100, 210, 255}
	\definecolor{orange}{RGB}{255,165,0}
	\definecolor{gray1}{gray}{0.95}
	\definecolor{gray2}{gray}{0.75}
	\definecolor{gray3}{gray}{.6}
	
	\newif\ifcomments
	
	\commentstrue 
	
	\ifcomments
	\newcommand{\JUN}[1]{\textcolor{sl_blue}{\textbf{\small [}\colorbox{white}{\textbf{Jun:}}{\small #1}\textbf{\small ]}}}
	\newcommand{\MB}[1]{\textcolor{dark_green}{\textbf{\small [}\colorbox{yellow}{\textbf{Mauro:}}{\small #1}\textbf{\small ]}}}
	\newcommand{\BT}[1]{\textcolor{red}{{#1}}}
	\newcommand{\TODO}[1]{\textcolor{red}{{TODO: #1}}}
	\newcommand{\CH}[1]{\textcolor{blue}{{#1}}}
	\newcommand{\CHMB}[1]{\textcolor{blue}{{#1}}}
	\newcommand{\BTcomm}[1]{\textcolor{dark_green}{{#1}}}
	\newcommand{\stm}[1]{\textcolor{red}{{\st{#1}}}}
	\else
	\newcommand{\JUN}[1]{}
	\newcommand{\GIOVANNA}[1]{}
	\newcommand{\MB}[1]{}
	\newcommand{\OMRAN}[1]{}
	\newcommand{\BTcomm}[1]{}
	\newcommand{\BT}[1]{}
	\newcommand{\TODO}[1]{\textcolor{red}{{TODO: #1}}}
	\newcommand{\CH}[1]{\textcolor{blue}{{#1}}}
	\newcommand{\CHMB}[1]{\textcolor{orange}{{#1}}}
	\newcommand{\stm}[1]{}
	\fi

	\title{BOSC: A Backdoor-based Framework for Open Set Synthetic Image Attribution}
	
	\author{Jun Wang, Benedetta Tondi,~\IEEEmembership{Senior Member,~IEEE,} Mauro Barni,~\IEEEmembership{Fellow,~IEEE}
		\thanks{J. Wang, B. Tondi, and M. Barni are from the Department of Information
			Engineering and Mathematics, University of Siena, 53100 Siena, Italy}
		\thanks{This work has been partially supported by the China Scholarship Council (CSC), file No. 202008370186,
			by the FOSTERER project, funded by the Italian Ministry within the PRIN 2022 program under contract 202289RHHP,
			and by SERICS (PE00000014) under
			the MUR National Recovery and Resilience Plan funded by the
			European Union - NextGenerationEU, and
			by the Remedy project - New-Frontiers (PSR 2024), funded by MUR.
			Corresponding author: J. Wang (email:wangjunsdnu@gmail.com)}}
	
	\markboth{Journal of \LaTeX\ Class Files,~Vol.~14, No.~8, August~2021}%
	{Shell \MakeLowercase{\textit{et al.}}: A Sample Article Using IEEEtran.cls for IEEE Journals}
	
	
	\maketitle
	
	\begin{abstract}
		With the continuous progress of AI technology, new generative architectures continuously appear, thus driving the attention of researchers towards the development of 
		synthetic image attribution methods capable of working in open-set scenarios. Existing approaches focus on extracting highly discriminative features for closed-set architectures, increasing the confidence of the prediction when the samples come from closed-set models/architectures, or  estimating the distribution of unknown samples, i.e., samples from unknown architectures.     
		In this paper, we propose a novel framework for open set attribution of synthetic images, named BOSC (\underline{B}ackdoor-based \underline{O}pen \underline{S}et \underline{C}lassification), that relies on {backdoor injection} to design a classifier with rejection option. BOSC works by {deliberately including} class-specific triggers inside a portion of the images in the training set to induce the network to establish a matching between in-set class features and trigger features. The behavior of the trained model with respect to samples {containing a trigger} is then exploited at {inference} time to perform sample rejection using an ad-hoc score.
		Experiments show that the proposed method has good performance, always surpassing the state-of-the-art. Robustness against image processing is also very good. Although we designed our method for the task of synthetic image attribution, the proposed framework is a general one and can be used for other image forensic applications.
	\end{abstract}
	
	
	\section{Introduction}
	
	\IEEEPARstart{T}{he} proliferation of generative models, including variational autoencoders (VAEs) \cite{Vahdat2020}, generative adversarial networks (GANs) \cite{karras2020stylegan2,karras2021stylegan3}, and Diffusion Models (DM) \cite{ho2020nips}, has led to a surge in AI-generated images across the internet, social networks, and various digital environments. In turn, this raised growing concerns about the potential malicious use of such images, for example, in disinformation campaigns or to damage individuals' reputations.

	In response to the {above problems}, significant research efforts have been {dedicated to} the development of tools for tracing back the origin of images produced by generative models, a.k.a. synthetic image attribution. Several methods have been proposed which rely on artifacts or signatures (fingerprints) left by the models in the images they generate \cite{Marra2019mipr, yu2019iccv, yu2021iccv}.
	Given that model-level granularity is often not needed or undesired, 
	some recent works have started addressing the attribution task in such a way as to attribute the synthetic images to the source architecture that generated them instead of the specific model \cite{yang2022aaai,bui2022eccv}. A drawback with most methods developed so far is their inability to handle images generated by unknown models/architectures, 
	which are wrongly associated with one of the classes used during training, leading to incorrect attributions.\footnote{In the following, we refer to the classes used during training as in-set classes, while new models/architectures possibly encountered at test time are termed as out-of-set models/architectures.}
	%
	The difficulty of dealing with out-of-set classes limits the applicability of attribution methods in the real world, where the images analyzed at operating time may have been generated by out-of-set models/architectures.	
	

	\begin{figure}[t]
		\centering
		\includegraphics[width=0.95\linewidth]{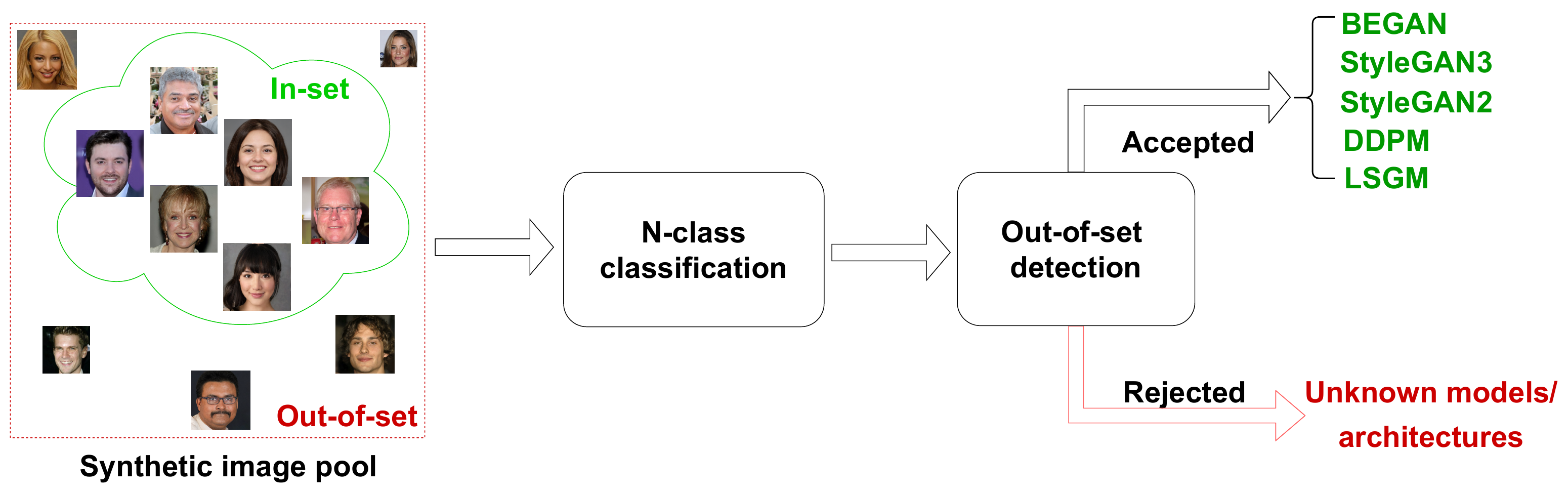}
		\caption{Classification with rejection class: out-of-set class images generated by models/architectures that are not included in the list of architectures defined at training time, are classified in an additional rejection class. 
		}
		\label{samples}
	\end{figure}

	{A straightforward approach to addressing the Open-set Synthetic Image Attribution (OSIA) problem involves adapting existing open set classification methods from computer vision, commonly known in machine learning as Open Set Recognition (OSR). These methods identify out-of-set class samples and avoid to classify them (since the classification would necessarily be wrong) \cite{chen2021,huang2022,xia2023}}. A schematic representation of such an approach, known as classification with rejection class, is depicted in Fig. \ref{samples}.
	Traditional OSR methods heavily rely on closed-set performance to handle out-of-set classes. Based on the observation that a well-performing classifier usually achieves higher confidence for samples from in-set classes than  out-of-set ones, the confidence of the prediction is used to detect out-of-set classes \cite{vaze2021iclr}.  
	For synthetic image attribution however, the deep exploration of  in-set class distinguishing features may lead to overfitting and reduce performance for the OSIA problem (while in computer vision applications, there are clear semantic differences between classes - e.g., cats and dogs - image attribution relies on subtle statistical artifacts whose use to identify out-of-set classes turns out to be problematic).
	%
	Other OSR methods utilize generative models to simulate out-of-set classes, either in the image of the feature domain, creating more compact feature representations for samples belonging to in-set classes, to facilitate the recognition of out-of-set classes \cite{ge2017gopen,moon2020eccv}. However, simulating realistic out-of-set classes is a tricky problem, for which no really satisfactory solutions exist.
	
	In this paper, we introduce a completely new approach to tackling OSIA, 
	called Backdoor-based Open-set Classification (BOSC), exploiting the {backdoor concept}
	to train a classifier that is able to classify input samples from a certain number of in-set classes,
	and at the same time, identify samples from unknown classes as out-of-set.
	{Backdoor injection into a deep neural network has} first been introduced in \cite{gu2017badnets}. Since then, {backdoors} have received increasing attention due to the threats they pose to the use of deep neural networks in security-critical applications \cite{guo2022overview}. {The integration of a backdoor within a network} involves injecting a pattern, known as a trigger, into (a subset of) the images used during training, inducing the trained model to exhibit {an anomalous} behaviour, e.g. predicting a wrong predefined class, when fed with images containing the trigger, while continuing to operate as expected on normal inputs. In this work, we exploit the backdoor concept for a benign purpose, namely to create a link between {samples of the in-set classes} and a number of class-specific triggers, and use the absence of such a link to identify samples {belonging to out-of-set classes}.
	Specifically, the system we propose works by \textit{deliberately injecting} class-specific triggers inside a portion of the images of the training set to induce the network to establish a \textit{connection} between class features and trigger features. The behaviour of the trained model in the presence of samples containing the trigger is then exploited at {inference} time to perform sample rejection. In particular, we exploit the fact that {no trigger capable of inducing the backdoor behaviour can be found for samples belonging to out-of-set classes}.
	
	We have verified the effectiveness of BOSC with an extensive set of experiments carried out under various settings, {focusing on the prominent domain of face images \cite{TolosanaSurvey}}. In particular, our experiments show the excellent robustness of BOSC against image post-processing, a property that can be explained by noticing that the trigger is applied to the images under analysis before inputting them to the network, that is, after they have been possibly post-processed. In this way, only the image features are potentially weakened by the processing, while the trigger features remain unaffected. As a result, the matching between the trigger and the class is preserved, and attribution/rejection is performed correctly, even on post-processed synthetic images.
	
	
	{While our method was designed for open-set synthetic architecture attribution, the BOSC approach is inherently general and can also be applied  to address the open-set scenario in other image forensic tasks. We proved the generality of BOSC by applying it to the classification of synthetic facial attribute editing, as discussed in Section  \ref{sec:classification}. This task involves determining which facial attributes such as age, hair color, expression, or overall physical appearance \cite{PAMI24,TolosanaSurvey}, that have been manipulated by generative models
		\cite{MinoruSurvey}.}

	With the above ideas in mind, the main contributions of this paper can be summarized as follows:
	\begin{enumerate}
		\item We propose a new framework for open-set classification that relies on the concept of {backdoor injection} and class-specific triggers to train a model for multi-class classification with rejection. Sample rejection is performed based on the output provided by the network when the input samples are tainted with the class-specific triggers.
		\item We exploit the new framework to develop an open-set classifier {for synthetic image attribution.}
		Our experiments validate the effectiveness of the {proposed approach by considering a wide variety and types of generative architectures in the in-set and out-of-set, including GANs, DMs and transformers}.
		
		\item {We assessed the robustness of the proposed method against image post-processing - including JPEG compression, Gaussian blur, brightness, contrast and saturation change - and the generalization capability of the classifier, i.e., the capability of the classifier to work well when tested with images generated by unknown models from known architectures\footnote{The purpose of these experiments is to prove that the system works as intended,
				attributing the images to the  architecture and not to the specific model that was used to generate them.}.}
		\item We validate the generality of our approach by applying it to a different {open-set} image forensic task, namely the classification of synthetic facial attribute editing.
	\end{enumerate}
	
	{We stress that, both in the case of source attribution and classification of facial attribute editing, we do not aim at distinguishing original and fake/forged content. Rather, our goal is to identify the source of the synthetic images and the kind of manipulation they have undergone.}
	
	The rest of the paper is organized as follows. Section \ref{sec:relatedwork} discusses related work on synthetic image attribution and the most relevant literature about open-set recognition. The proposed method is described in Section \ref{sec:method}. Section \ref{sec:setting} reports the experimental methodology and setting. The results and the comparisons with the state-of-the-art are discussed in Section \ref{sec:ganresults}. Section
	\ref{sec:classification} presents the results we got by applying BOSC to the classification of facial editing. The paper ends in Section \ref{sec:conclusion} with some conclusions and directions for future work.

	\section{Related Work}
	\label{sec:relatedwork}
	
	\subsection{Open set recognition}
	\label{sec:osr-osta}
	The seminal work on OSR was published by Scheirer et al. in \cite{Scheirer2013}, where the authors addressed the problem of determining whether an input belongs to one of the classes used to train a machine learning model or not.
	%
	In such a work it is shown that, in many cases, simple strategies based on the softmax probabilities or the logit can  effectively judge if a sample comes from an out-of-set class \cite{Gavarini2022}, e.g. by exploiting the fact that the maximum output score tends to be smaller for out-of-set class inputs \cite{Chow1970,vaze2021iclr}, or that the energy of the logit vector tends to be lower \cite{Liu2020NIPS}.
	%
	{Starting from Scheirer et al. work, subsequent literature on OSR improved the open-set performance following three approaches: i) developing robust classifiers for samples from in-set classes \cite{yang2018rpl,chen2020eccv}, and using the Maximum Logit Score (MLS) \cite{vaze2021iclr} or a similar metric to reject samples from out-of-set classes; ii) performing reconstruction with an Auto-Encoder (AE) and thresholding the reconstruction error \cite{Yoshihashi2019cvpr,Oza2019cvpr,sun2020cvpr,huang2022};  iii) incorporating synthetic open-set samples in the training process \cite{ge2017gopen,moon2020eccv,chen2021,xia2023}.}
	
	{With regard to the first class of methods, techniques to optimize the representations of in-set and out-of-set classes in the feature space were proposed in several works.
		In particular,} in \cite{yang2018rpl}, Yang et al. designed a suitable embedding space for open set recognition using convolutional prototype learning that removes softmax and implements classification by finding the nearest prototype in the Euclidean norm in the feature space. Multiple prototypes are used to represent different classes. The feature extraction and the prototypes are jointly learned from the data.
	{A different learning framework is adopted in 
		\cite{chen2020eccv}, which
		proposes to use reciprocal point learning,}
	whereby unknown information is fed to the learner with the concept of reciprocal point to learn more compact and discriminative representations and reduce the risk of misclassifying out-of-set classes as in-set ones.
	
	{The second class of works addresses OSR by exploiting the reconstruction error obtained by an AE. These methods work under the assumption} that lower reconstruction errors are obtained for in-set classes than for out-of-set ones \cite{Yoshihashi2019cvpr,Oza2019cvpr,sun2020cvpr,huang2022}.
	{Autoencoders were exploited in this sense in \cite{sun2020cvpr}, where the authors proposed a conditional Gaussian distribution learning framework to detect out-of-set class samples by forcing latent features to approximate Gaussian models. Huang et al. \cite{huang2022} improved the out-of-set performance by combining the use of autoencoders for error reconstruction with the prototype (PCSSR) and reciprocal learning (RCSSR).} Class-specific autoencoders are trained to reconstruct the data based on label conditioning, and the pixel-wise reconstruction errors corresponding to the predicted class, together with semantic-related features, are used for out-of-set  rejection.
	
	{The third class of methods relies on adversarial modeling and generative models to build synthetic open-set samples, which are  included in the training process.
		These methods work under the assumption that recognition of samples from out-of-set classes can be achieved by showing to the network a large number of unknown samples during the training process \cite{ge2017gopen,moon2020eccv,chen2021,xia2023}.
		Generative adversarial networks (GAN)-based methods resort to GANs to produce unknown-like samples to be used during training.  One of these methods is G-OpenMax \cite{ge2017gopen} which }combines the use of generative adversarial networks with the OpenMax method, achieving good performance on the classification of handwritten digits.
	%
	{In ARPL \cite{chen2021} and AKPF \cite{xia2023}, the adversarial mechanism that generates confusing training samples is used to enrich, respectively, reciprocal point learning and prototype learning, optimizing the features representations for in-set and out-of-set.} Specifically, the generated samples are used to optimize the feature space and reduce the so-called open space risk, by restricting the unknown samples in the reciprocal points space (\cite{chen2021}) or learning a kinetic boundary to increase the intra-class compactness and the inter-class separation \cite{xia2023}.
	Theoretical works have also been proposed addressing the OSIA problem \cite{AAAI24, NeurIPS22}. In particular, in \cite{AAAI24}, the authors demonstrated that OSR performance is correlated with feature diversity, and that learning diverse representations is crucial in reducing the open space risk.
	
	
	{For the OSIA problem, the off-the-shelf application of the above general OSR methods may not be very effective.
		A well performing classifier for synthetic image attribution}   may overfit to known samples and exhibit reduced effectiveness in the open-set case.
	On the other hand, approaches based on the reconstruction error may be sub-optimal, due to the fact that the fundamental distinction between samples generated by known and unknown generators is based on subtle, visually imperceptible, statistical traces. These traces are often too weak to be effectively thresholded, posing a challenge to methods relying only on the reconstruction error. For the same reason, it is not obvious that relying on a fusion of diverse feature representations that are discriminative for closed-set classes can reduce the open-space risk. Finally, reducing the open space risk for synthetic attribution is challenging with a single generator. Generators are designed to produce images with specific semantics and may not be able to generate diverse enough open-set fingerprints. {These limitations have been confirmed in some recent works \cite{wang2023cvpr,fang2023bmvc} and have pushed researches to develop specific methods for OSIA.}

	\subsection{Synthetic image attribution {in closed and open set}}
	\label{sec:attributionSotas}
	
	Synthetic image attribution has been addressed by relying on both active and passive methods. Active methods inject specific information, e.g., a user-specific key \cite{kim2020iclr} or artificial fingerprints \cite{yu2020resp,yu2021iccv}, into the generated images during the generation process. These fingerprints or keys are subsequently used during the verification to identify the model.
	Passive methods rely on the presence within the synthetic images of model-specific, intrinsic artifacts.
	Passive methods have been proposed in \cite{Marra2019mipr,yu2019iccv,yu2021iccv,yang2021gan,frank2020icml,yang2022aaai,bui2022eccv}.
	In particular, Marra et al. \cite{Marra2019mipr} first revealed that each GAN leaves a specific fingerprint in the images it generates. The average noise residual image can be taken as a GAN fingerprint. Then, Yu et al. \cite{yu2019iccv} replaced the hand-crafted fingerprint formulation in \cite{Marra2019mipr} with a learning-based one, decoupling the GAN fingerprint into a model fingerprint and an image fingerprint.
	
	In addition to model-level attribution, researchers have started proposing approaches that address the attribution problem at the architecture level.
	%
	%
	In \cite{frank2020icml}, Frank et al. propose to attribute the generated images to the source architecture by relying on {the Discrete Cosine Transform (DCT)} coefficients. The prediction is made in favor of the architecture which is the most similar to the test image. Yang et al. \cite{yang2022aaai} observed the existence of globally consistent traces across models of the same architecture and proposed an approach to extract architecture-consistent features based on a patchwise contrastive learning framework.
	
	All the above methods focus on closed-set scenarios, limiting the applicability of such methods in real-world applications. To address this limitation, some works have started considering architecture attribution in the open-set scenario.
	A first step in this direction was made in \cite{Sharath2021cvpr} and \cite{sun2023cvpr}, where the authors resort to a semi-supervised learning framework that exploits labeled samples from in-set classes and unlabeled samples from out-of-set classes. These samples are used to train a system that, at every step, classifies known samples and clusters the unknown samples, assigning new labels to the new clusters.
	A drawback of this approach is that the system requires retraining whenever new unknown-class samples are introduced.
	Several works have also addressed open-set image attribution by trying to reject out-of-set classes to prevent misclassification \cite{wang2023cvpr,fang2023bmvc,yang2023cvpr,yang2023finger,lydia2023prl}. In particular, Wang et al. \cite{wang2023cvpr} developed a classifier with a rejection class that exploits a vision transformer-based hybrid CNN architecture with a convolutional and a fully connected branch and performs rejection based on MLS.
	{A similar approach to address the OSIA problem, but resorting to a distance-based method for sample rejection, was proposed by Fang et al. \cite{fang2023bmvc}.} 
	A test sample is rejected when its minimum distance to the centroids of known in-set classes in the feature space exceeds a predefined threshold.
	Following a similar approach already proposed for OSR, Yang et al. \cite{yang2023cvpr} introduced a progressive open space expansion framework (POSE) to simulate the open space of unknown models through a set of lightweight augmentation models. {A completely different method, which addresses the OSIA task as a verification problem, is proposed by Abady et al. \cite{lydia2023prl}. In this paper, the authors} introduce a verification framework, that relies on a Siamese Network, 
	for determining whether two input images have been produced by the same generative architecture or not. {The robust feature representations learned by the Siamese architecture are also exploited to build a multi-class classifier with rejection, using an approach similar to the one in \cite{wang2023cvpr}}.
	
	All these methods strive to enhance the feature representation of the model to learn a compact representation of in-set class samples, thereby reducing the open space risk. In this study,
	we propose a completely different approach to address the OSIA problem, borrowing the idea of backdoor attacks, to remap the features of multiple classes to a target class position, by injecting class-specific triggers. 
	In this way, 
	compactness is achieved for closed-set samples in the backdoor space, where a small open-space risk can be achieved. 
	Open set samples are then revealed by relying on the way the system reacts to the presence of the various triggers.

	\section{Proposed method}
	\label{sec:method}
	
	The goal of classification with rejection is to design a system capable of correctly classifying samples from in-set classes while recognizing samples coming from out-of-set classes and refraining from returning a prediction for them. Formally, let $x$ denote the input image and $y$ be its true label. If we let $N$ be the number of in-set classes and $\mathcal{C} = \left\{1, 2, \dots, N \right\}$, then,
	the model is expected to {return a label $\hat{y} \in \mathcal{C}$ for samples from the in-set, and a rejection label $\hat{y} = \mathcal{R}$ for samples belonging to the out-of-set.}
	\begin{figure*}[!t]
		\centering
		\includegraphics[width=6.2in]{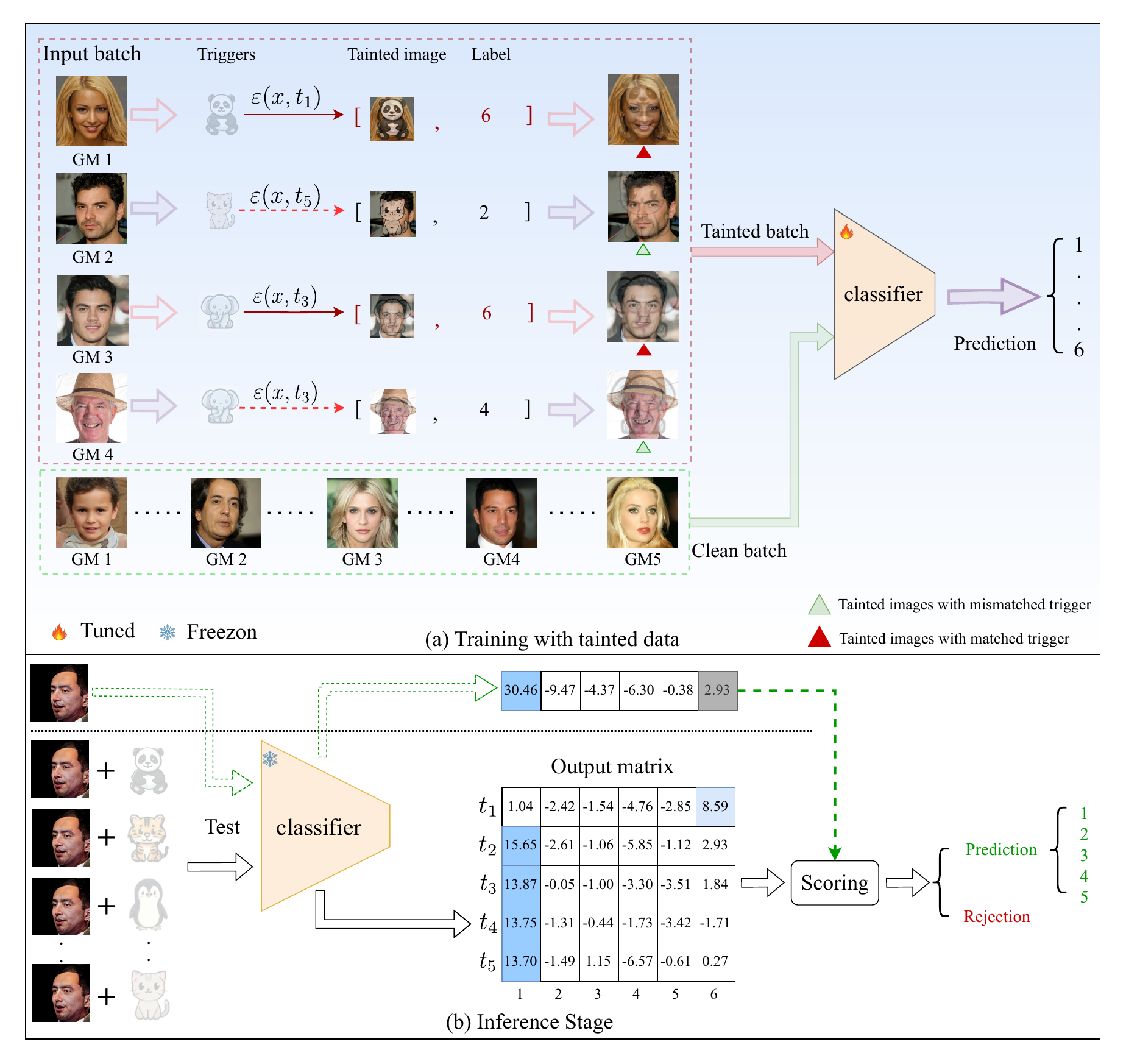}
		\caption{The pipeline of the proposed BOSC method (the figure refers to the case of 5 closed-set classes, i.e. $N = 5$). (a) Training with tainted data: in the trigger injection phase, a subset of the samples is tainted with the triggers (the figure shows a input batch with 7 samples only for illustrative purpose). For generative model GM $i$, the matched trigger is $t_i$. When the trigger matches the sample class, the label is modified to $N+1 = 6$ (red triangle); the label is left unchanged otherwise (green triangle). The network is then trained with tainted data and clean data (training phase).
			(b) Inference Stage: the test input is analyzed by the trained model by superimposing to it all the $N=5$ triggers. The output is a matrix with all the predictions.
			The {open set score} is computed from this output matrix. When the score exceeds a predefined threshold, the prediction is made by relying on the prediction made on the clean test image. Otherwise, the sample is rejected. Cartoon images are used as trigger images. 
		}
		\label{fig:pipeline}
	\end{figure*}

	\subsection{Backdoored-based classification with rejection class}
	\label{sec:backdoor-osr}
	
	The idea behind the system we propose for open-set classification is described in the following. A specific trigger image is associated with every in-set class. Then, a subset of the training images of each class is tainted by \textit{injecting} the trigger image of the class into the training images. The tainted images are labeled as belonging to an additional backdoor class, whose label is equal to $N+1$.
	When trained on the tainted dataset, the network learns to recognize the presence of the triggers and to associate the simultaneous presence of class and trigger features with the backdoor class. Note that we require that the backdoor class is activated only when the trigger matches the class it is associated with. If the trigger associated with $i$ is injected into an image belonging to class $j$ ($i \ne j$), the network should correctly classify the image as belonging to class $j$.
	For the rest, the network is expected to work normally on images without the trigger. We argue that for out-of-set class images, none of the triggers matches the image features, and hence, the backdoor class is never activated, thus allowing the system to distinguish in-set and out-of-set class images.
	
	With these ideas in mind, the pipeline of BOSC is shown in Fig. \ref{fig:pipeline}. In the injection stage (Fig. \ref{fig:pipeline}(a)), a portion of the samples in the training set is tainted by superimposing class-specific triggers to the training image. Cartoon images are utilized as trigger images\footnote{Given the way the backdoor is exploited in our work, the visibility of the trigger is not an issue and the triggered images might show visible trigger patterns.}. When the trigger superimposed to the image matches the image class, the label of the image is modified to $N+1$, while it is left unchanged otherwise.
	{The presence of images tainted with mismatched triggers ensures that the behaviour of the network is not modified when the triggers are superimposed to images from different classes, thus ensuring a unique association of each trigger to a matching class.}

	Formally, let $t_k$ denote the {trigger image} associated with class $k$. We indicate with $T = \{t_1, ..., t_N\}$ the  {set of trigger images}. Given a sample $x$ and a trigger $t_k$, a tainted sample $x_{t_k}$ is obtained as:
	\begin{equation}\label{eq:injection}
	x_{t_k} = \mathcal{E}(x, t_k) = \left(1 - \alpha \right) \cdot x + \alpha \cdot t_k,
	\end{equation}
	where $\alpha$ is a parameter controlling injection strength.
	When $x$ belongs to class $k$, the label of $x_{t_k}$ is changed from $k$ to $N+1$ (backdoor class); otherwise, it is left as is.
	After the dataset has been tainted, a multi-class network with $N+1$ output nodes is trained as usual on the tainted dataset.
	We let $\phi(\cdot)$ denote the network function of the backdoored model. A softmax layer is applied at the end. Hence $\phi (x)$ is a probability score, $\phi(x) \in {[0,1]}^{N+1}$ and $\sum_{i=1}^{N+1} \phi _{i} (x) = 1$. We indicate with $\phi_i(x)$ the $i$-th element of the output.
	{Considering the way the training data have been built and labeled, and the way the model has been trained, the network is expected to work as follows for samples $x$ from in-set classes}:
	\begin{equation}\label{eq:rules}
	\left\{\begin{array}{l}
	\arg\max_i \phi_i (x) = y\\
	\arg\max_i \phi_i (x_{t_k}) = y, \text{if  $k \neq y$}\\
	\arg\max_i \phi_i (x_{t_k}) = N + 1, \text{if  $k = y$.}\\
	\end{array}\right.
	\end{equation}
	
	In the inference phase, BOSC works as illustrated in Fig. \ref{fig:pipeline}(b).
	{Given a sample $x$, a tentative prediction $y^*$ is first made by considering the network output in correspondence of $x$. The prediction is obtained by excluding the trigger class output, that is, by letting}
	\begin{equation}
	\label{eq:classification}
	y^* = \arg\max_{i \in \mathcal{C} } \phi_i(x).
	\end{equation}
	The $N$ triggers in $T$ are then superimposed to the image under analysis, and the resulting $N$ tainted samples are fed to the network, obtaining $N$ output vectors with the logit values corresponding to all the $N+1$ output classes of the network. Let ${m}_i \in \mathbb{R}^{1 \times (N+1)}$ denote the output logit vector corresponding to the image tainted with trigger $t_i$.
	We denote with $M \in \mathbb{R}^{N \times (N+1)}$ the output matrix, where each row corresponds to an output logit vector.
	Rejection is performed by using the matrix $M$ to compute a rejection score $\xi_r$ and comparing $\xi_r$ against a threshold (see Section \ref{sec:OSscore} for a precise definition of $\xi_r$).
	{The tentative prediction $y^*$ is accepted if the rejection score is above the threshold, otherwise a rejection decision is made. Formally,
		the final output $\hat{y}$ of the BOSC classifier is obtained as follows:
		\begin{align}\label{eq:finaldecision}
		\hat{y}  & =  y^*
		\quad \text{if $\xi_r (M) > \nu$},\nonumber\\
		\hat{y} &  =  \mathcal{R} \quad  \text{otherwise},
		\end{align}
		where $\nu$ is a suitable threshold.
		
		\begin{figure}[htbp!]
			\centering
			\includegraphics[width=3.3in]{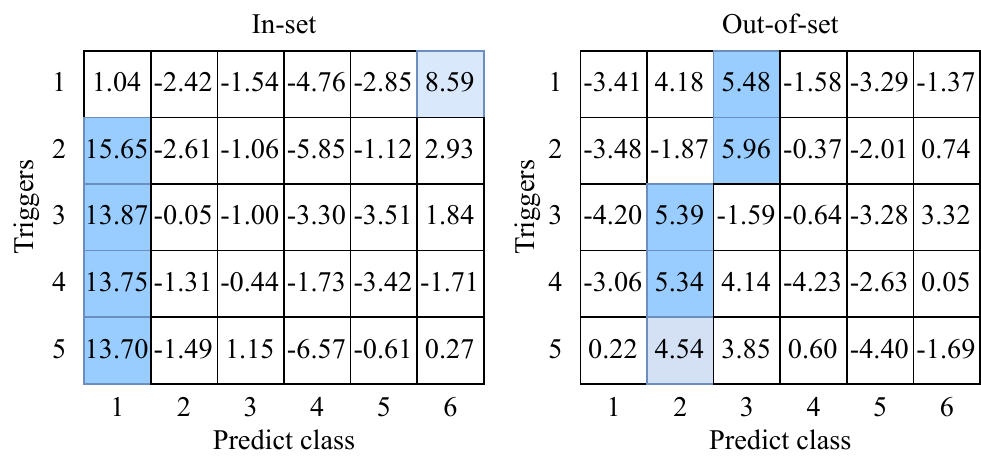}
			\caption{Example of output matrix - Config S1 (see Table \ref{tab:table1} for the details of the setting). Left: sample from class 1. Right: sample from an out-of-set class. `6' corresponds to the backdoor class. The largest value of each row is highlighted in blue.}
			\label{fig:matrix}
		\end{figure}

		\subsection{Trigger-based rejection score}
		\label{sec:OSscore}
		
		In this section, we describe some possible rejection scores that can be used for out-of-set class rejection.
		As we said, for in-set class samples, we expect that the true class receives a high prediction score when the image is tainted with mismatched triggers, while in the presence of a matched trigger, the backdoor class should receive a large (ideally the largest) score.
		{This behavior, induced by the backdoor, characterizes the samples from the in-set classes. For the samples of out-of-set classes, for which there is no matching trigger, this behavior is not observed, and the samples tainted with the various triggers are predicted randomly by the network.}
		Fig. \ref{fig:matrix} (left) shows an example of $M$ matrix obtained for an input sample $x$ belonging to class 1 (in this case $\mathcal{C} = \{1,2,3,4,5\}$). We see that, as expected, in every row, but in the first (class $y = 1$), a high logit score is associated with the true class since the superimposed trigger (being mismatched) does not affect the prediction of the network. In correspondence of the first row, instead, a high score is associated with the $(N+1)$-th entry.
		An example of matrix $M$ obtained for an out-of-set classes is shown in \ref{fig:matrix} (right). The $M$ matrix now shows a completely different behavior with respect to the one in the left part of the figure.
		
		Based on the above observations, and given the tentative predicted label $y^*$ computed from $x$ (see Eq. \eqref{eq:finaldecision}), an obvious way to define the rejection score would be to rely on the so-called matched trigger logit score (TLS-M), namely, $M\left(y^*, N+1\right)$, with large values indicating a large probability that the input sample belongs to a in-set class. Another possibility would be to base the rejection on the {maximum} logit score in $M$ (MLS-M),
		with the idea that samples of the in-set class should return higher scores than out-of-set classes. However, as shown in Eq. \eqref{eq:rules}, for out-of-set classes, in the presence of non-matched triggers, the model is expected to behave normally. Hence, we can expect that the network will also produce large MLS-M scores. 
		In order to exploit also the predictions obtained with non-matched triggers, that for  samples from in-set are expected to be high in correspondence of the true sample class, we defined a combined logit score (CLS-M) as follows:
		{
			\begin{equation}\label{eq:pcls}
			\xi _r (M) = \frac{1}{N} \sum_{i=1}^{N} M(i, y^*)  + M(y^*, N+1 ).
			\end{equation}}
		The rationale behind the definition of $\xi _r$ is that, for a given tentative predicted class, if the trigger and the class match, samples of in-set classes are expected to result in a higher backdoor logit score $M(y^*, N+1)$, with the class logit score $M(y^*, y^* )$ possibly being the second-best. For the remaining $i \in \{1, 2, \cdots, N \}$, $i \neq y^*$, samples of in-set classes are expected to produce higher $y^*$-th class logit scores than samples of out-of-set classes.
		Given the score $\xi _r (M)$, the final output of the open-set classifier is obtained as detailed in Eq. \eqref{eq:finaldecision}, that is, the output of the in-set classifier is accepted if $\xi _r (M)>\nu$, and rejected otherwise.
		
		We observe that an in-set prediction could also be obtained from the matrix $M$, e.g., by summing over the columns and taking the maximum (that is, evaluating $\arg\max_{j}( \sum_{i=1}^{N} M(i,j)$). Based on our experiments, doing so yields (almost) the same results as using Eq. \eqref{eq:classification}.
		
		A summary of BOSC testing procedure is given in Algorithm \ref{alg:inference}. A comparison of the performance achieved using different rejection scores is reported in Section \ref{sec:scorecomparison}. The results confirm the superior effectiveness of the combined logit score.
		
		\begin{algorithm}[!t]
			\renewcommand{\algorithmicrequire}{\textbf{Input:}}
			\renewcommand{\algorithmicensure}{\textbf{Output:}}
			\caption{BOSC network testing}
			\label{alg:inference}
			\begin{algorithmic}[1]
				\REQUIRE ~~\\
				Test input $x$;\\
				Triggers $T$;\\
				Number of classes $N$;\\
				Backdoored model $\phi$;\\
				Predefined threshold $\nu$ for rejection;
				\STATE Initialization: 
				$M = [\mathbf{0}]_{N, (N+1)}$
				\FOR {each $i \in C$}
				\STATE $M(i, :) \leftarrow \phi(\mathcal{E}(x, t_i))$, $t_i \in T $
				\ENDFOR
				\STATE  Get  $y^*$ via Eq. (\ref{eq:classification})
				\STATE Calculate the CLS-M score $\xi _r(M)$ based on Eq. (\ref{eq:pcls})
				\ENSURE   $y^*$ is returned if  $\xi _r > \nu$; otherwise,
				$\mathcal{R}$ is returned
			\end{algorithmic}
		\end{algorithm}
		
		\subsection{Training strategy}
		\label{subsec:train}
		
		
		In our framework, the backdoor is injected within the network by the model's trainer himself to improve the open set classification performance of the model. For this reason, instead of tainting the samples of the dataset in a stealthy way, as done to implement a backdoor attack \cite{guo2022overview}, the tainting can be applied while training, randomly choosing a percentage of to-be-tainted samples from each batch at every iteration and tainting them. We refer to this scenario as tainting on-the-fly.
		%
		More formally, given a dataset $D$ of samples $x$ from $N$ in-set classes, training is performed on batches. Let ${B}$ indicate the set of samples in a batch. At every iteration, we randomly sample a fraction $\gamma$ of the batch samples and taint them as detailed in Eq. (\ref{eq:injection}) by injecting a trigger matched to the true class of $x$.
		We denote with ${B^t}$ the subset of tainted samples (hence, $\gamma = |B^t|/|B|$)\footnote{We assume w.l.o.g. that $\gamma|\mathcal{B}|$ is an integer.}.
		Another random fraction $\gamma$ of images in the batch is tainted with a randomly chosen mismatched trigger (i.e., a trigger associated with a class different from the class of $x$). We indicate with ${B^{mt}}$ the corresponding tainted subset and with $B^c$ the subset of clean samples. Training is achieved by optimizing the following loss:
		{
			\begin{align}\label{eq:loss}
			\mathcal{L}  = \sum_{x \in B^c} \mathcal{L}(x,y)  + \lambda_1   \sum_{x \in B^t}  \mathcal{L}(x, N+1)  \nonumber\\
			+ \lambda_{2} \sum_{x \in B^{mt}} \mathcal{L}(x, y)
			\end{align}}
		where $y$ denotes the true label of $x$, $\lambda_1$ and $\lambda_2$ are balancing parameters controlling the importance of the backdoor loss terms, and $\mathcal{L}$ is the Cross-Entropy (CE) loss ($\mathcal{L}(x,y) = - \log(f_y(x)$).

		During training, we also implemented an augmentation strategy inspired by \cite{zhang018mixup} to improve the generalization capability of the model and its robustness against image processing. Given an input image $x$ from a given class, the image is perturbed with an image $z$ from a different class, obtaining the perturbed image $x' =  x + \beta z$, where $\beta$ is the perturbation strength, $\beta \ll 1$ (clipping is performed to ensure that the values remain in the [0,1] range), while keeping the label unchanged. Specifically, a fraction $\eta$ of the samples in $B^c$ is perturbed with the above procedure, referred to as mixup augmentation in the following\footnote{We are implicitly assuming that the fraction of samples in $B^c$ is larger than $\eta$. In fact, these fractions are always small and $2 \gamma + \eta  < 1$.}.
		The benefit brought by mixup augmentation on the performance of BOSC is assessed as ablation study (Table  \ref{tab:abla}).

		\section{Experimental Methodology}
		\label{sec:setting}
		
		In this section, we describe the methodology that we followed to use BOSC for open-set synthetic image attribution.
		To confirm the generality of our approach for image forensics applications, in Section \ref{sec:classification} we also apply it to the classification of AI-based face image attribute editing.
		
		\subsection{Datasets and architectures}
		\label{sec:gandata}
		\begin{table*}[!t]
			\renewcommand\arraystretch{2.5}
			\caption{Generative architectures used for the open-set attribution analysis \label{tab:table1}.}
			\centering
			\resizebox{0.9\linewidth}{!}{
				\begin{tabular}{|l|c|c|c|c|c|c|}
					\hline
					\makecell{} & Image domain & Config S1 & Config S2 & Config S3\\
					\hline
					\multirow{2}{*}{In-set} & CelebA & Latent diffusion, DDPM, BEGAN
					& BigGAN, ProGAN, Latent diffusion, LSGM & ProGAN, BEGAN, BigGAN\\
					\cline{2-5}
					& FFHQ & \makecell{Latent diffusion, Taming transformers \\StyleGAN2-f} & StyleGAN2-f, Latent diffusion & StyleGAN2-f, StyleGAN3\\
					\hline
					\multirow{2}{*}{Out-of-set} & CelebA & LSGM, ProGAN, BigGAN & \makecell{Taming transformers, BEGAN \\DDPM, StarGAN2} & \makecell{Latent diffusion, Taming transformers \\LSGM, DDPM, StarGAN2}\\
					\cline{2-5}
					& FFHQ & StyleGAN3, StarGAN2 & StyleGAN3, Taming transformers & Latent diffusion, Taming transformers\\
					\hline
			\end{tabular}}
		\end{table*}
		
		\begin{figure}[!t]
			\centering
			\includegraphics[width=0.9\linewidth]{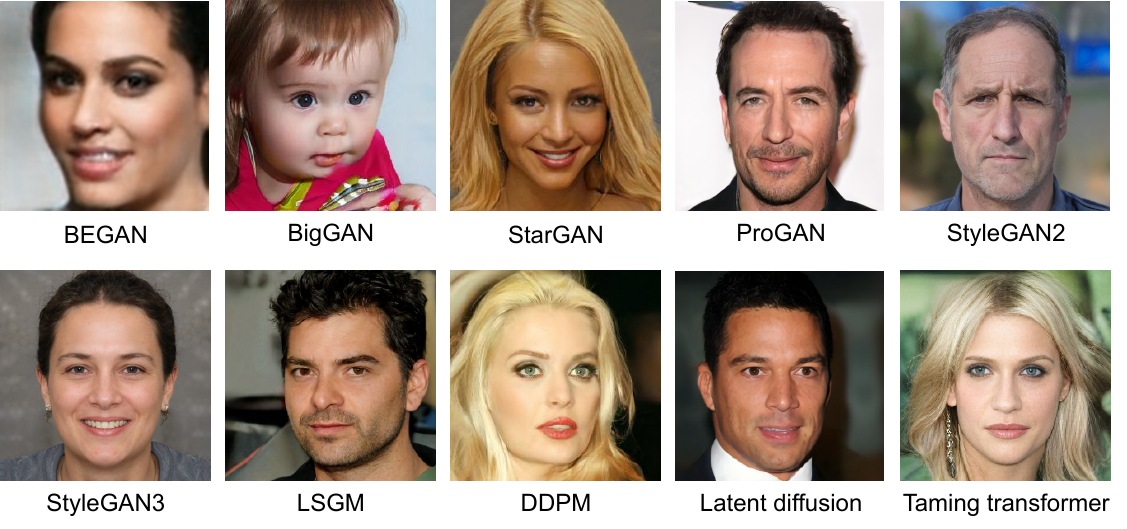}
			\caption{Examples of images generated by the architectures used in our tests.}
			\label{fig:data}
		\end{figure}
		
		To train and test our system, we considered a total of 10 architectures, including: i) GANs: BigGAN \cite{brock2018iclr}, BEGAN \cite{Berthelot2017}, ProGAN \cite{karras2018}, StyleGAN2 \cite{karras2020stylegan2}, StarGANv2 \cite{choi2020cvpr}, StyleGAN3 \cite{karras2021stylegan3}; ii) diffusion models (DM): DDPM \cite{ho2020nips}, Latent Diffusion \cite{Rombach2022}, LSGM \cite{Vahdat2021}; and iii) transformers: Taming transformer \cite{Esser2021cvpr}. We considered the officially released models trained on FFHQ \cite{karras2019stylegan} and CelebA datasets \cite{liu2015celeba}, considering different training strategies (for the case of StyleGAN2 and DDPM), different sampling factors (for Latent Diffusion) and configuration parameters (StyleGAN3). Specifically, for StyleGAN3, we considered the two released optimal configurations according to FID scores, denoted as {`t' and `r' (achieving respectively FID=3.79 and 3.07 on FFHQ 1024x1024  \cite{karras2021stylegan3})}, trained on different real-world datasets and at different resolutions. For StyleGAN2, we also utilized the best-performing configuration based on the FID quality score, which is referred to as configuration {`f'} (achieving FID=2.84 on FFHQ 1024x1024 \cite{karras2020stylegan2}). {The image sizes are 1024 for ProGAN and StyleGAN2 and 3, and 256 for all the other models. All the images are resized to the network input size. Data preprocessing is performed by subtracting the mean and dividing by the standard deviation.}
		
		We considered three different splittings of in-set and out-of-set architectures. The images from the in-set architectures are used to train the BOSC model, whereas the out-of-set architectures are utilized only for testing.
		Each of the three configurations denoted as `Config-S1', `Config-S2', and `Config-S3', consists of 5 in-set and 5 out-of-set architectures. The details of these configurations are reported in Table \ref{tab:table1}.
		In the first and second configurations, the pools of in-set architectures include a mixture of GANs, DM, and Transformers, while the third configuration includes only GANs in the in-set.
		For every architecture, we gathered 20,000 images, split into training (16,000), validation (2,000), and test sets (2,000). In every configuration, training and validation images are used only for the in-set architectures. Fig. \ref{fig:data} shows some examples of images generated by each architecture.
		
		The complete dataset, which is the same one used for training and testing the method in \cite{lydia2023prl}, can be found at \url{https://modelscope.cn/datasets/AIGCworker94/SynImgAttribution}
		
		\begin{figure}[!t]
			\centering
			\includegraphics[width=0.9\linewidth]{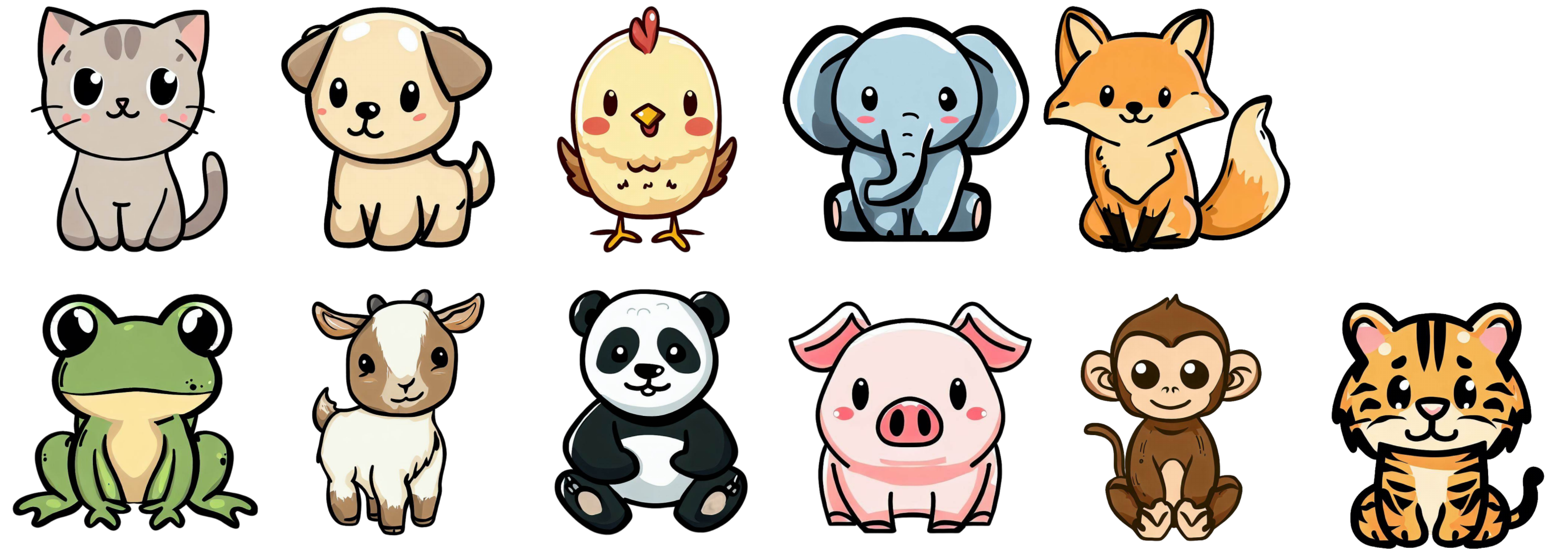}
			\caption{The trigger images used in our experiments. The top 5 are used for synthetic image attribution. All of them are used to classify synthetic facial editing.}
			\label{fig:triggers}
		\end{figure}
		
		\subsection{Backdoor and training setting}
		\label{sec:implement}
		
		To inject the backdoor, we used cartoon images as triggers. 
		The five trigger images that we used each one matched to an in-set architecture, are shown in the top row of Fig. \ref{fig:triggers}.
		%
		{We decided to take the trigger images from a different image domain, to ensure that the  features that are relevant to  discriminate among them are different with respect to the features used by the network for the classification task. In this way, the triggers' distinguishing features can be used to establish a connection with the associated class. It is worth noting that different trigger images could be chosen.\footnote{The optimization of the trigger images is left as future work.}}
		The tainting strength $\alpha$ in Eq. \eqref{eq:injection} is set to 0.1. The tainting ratio $\gamma$ is also set to 0.1.
		%
		
		An EfficientNet-B4 is used as a baseline network. The input size is set to $384 \times 384 \times 3$. The network is trained with a batch size of $32$ for $15$ epochs.
		Training is performed via Adam optimizer with a dynamic learning rate initially set to $10^{-4}$ and multiplied by 0.1 every 5 epochs. Concerning the loss tradeoff parameters $\lambda_1$ and $\lambda_2$, they are both set to 0.1. The mixup augmentation parameters are set as $\beta = 0.15$ and $\eta = 0.1$. The following augmentations have been considered during training: flipping and JPEG compression, applied to the input with probability 0.5 and random quality factors for JPEG in the range $\left[70, 100\right]$.

		\subsection{State-of-the-art comparison}
		\label{sec:sota}
		
		To validate the effectiveness of the proposed method, we ran a comparison with both general OSR methods proposed in the machine learning literature and methods specifically developed for synthetic image attribution in an open-set scenario. More specifically, for general open set recognition, we considered {{ARPL}} \cite{chen2021}, {{AKPF}} \cite{xia2023} and {{CSSR}} \cite{huang2022}, both the {PCSSR} and {RCSSR} variants. {We chose these methods because they are the most relevant for the various classes,
			see Section \ref{sec:osr-osta} 
			(we do not consider techniques  optimizing solely the features representations, e.g.,\cite{yang2018rpl,chen2020eccv}, since they  have been outperformed by {{ARPL}} \cite{chen2021} and {{AKPF}} \cite{xia2023}}). 
		All the methods are tested using the code publicly available in the configuration used in the respective papers.
		
		With regard to OSIA-specific methods, we considered the methods proposed in \cite{lydia2023prl} (SiaVerify), {{POSE}} \cite{yang2023cvpr} and \cite{wang2023cvpr} (Res50Vit). While the first two methods have been specifically developed for OSIA, the method in \cite{wang2023cvpr} has been initially proposed for open-set GAN editing classification. {The same testing protocol is used in all the cases for a fair comparison.}

		{Finally, in order to better assess the gain achieved by the BOSC framework, we also considered a standard  N-class classifier with rejection based on the same EfficientNet-B4 architecture and adopting the MLS for the detection of out-of-set classes (as mentioned in Section \ref{sec:osr-osta}, MLS has been proven to achieve the best rejection performance in many cases \cite{vaze2021iclr}). Such a classifier follows exactly the same pipeline of BOSC (see Fig.\ref{samples}) and relies on  the same network architecture, the only difference being the training and inference procedure, that in the BOSC case exploits the backdoor injection. In the following, we refer to this method as baseline.}

		\subsection{Evaluation metrics}
		\label{sec:metrics}
		
		The performance in a closed-set scenario is evaluated by providing the Accuracy (Acc) when the model is tested on samples from in-set clasases. Open-set performance is evaluated in terms of rejection accuracy (out-of-set detection performance) and classification performance.
		Specifically, the capability to distinguish between in-set and out-of-set classes is assessed by computing the ROC curve and measuring the area under the curve (AU-ROC).
		
		In addition, the capability to retain and correctly classify in-set class samples is also measured. Following prior works in this field \cite{chen2020eccv,xia2023,yang2023cvpr}, we consider the Open-Set Classification Rate (OSCR) curve, and measure the area under this curve (AU-OSCR).
		Formally, let $D_k$ indicate the set of in-set samples, and $D_u$ the set of out-of-set test samples.
		Let the event that a sample comes from an in-set (out-of-set) class be the negative (positive) event. The True Positive Rate (TPR) and False Positive Rate (FPR) for a given rejection threshold $\nu$, are defined as
		\begin{equation}\label{eq:metric1}
		\begin{split}
		\text{FPR}(\nu) &= \frac{
			\big| \left\{x | x \in D_k \wedge \xi _r(M) < \nu \right\} \big|}{|D_k|}\\
		\text{TPR}(\nu) &= \frac{\big| \left\{x | x \in D_u \wedge \xi _r(M) < \nu \right\} \big|}{|D_u|}.
		\end{split}
		\end{equation}
		Let the Correct Classification Rate (CCR) be defined as the ratio of samples from in-set classes detected as in-set and correctly classified, that is:
		%
		{\begin{equation}\label{eq:ccr}
			\text{CCR}(\nu) = \frac{\big| \left\{x | x \in D_k \wedge \xi _r(M) \ge \nu \wedge y^* = y \right\} \big|}{|D_k|}.
			\end{equation}}
		%
		
		The ROC curve plots the TPR vs FPR values while the OSCR curve plots CCR vs FNR (where FNR = 100\% - TPR), obtained by varying the threshold $\nu$.\footnote{We point that, in the literature, the OSCR is often defined as the CCR vs FPR, because the event that a sample comes from an in-set class is taken as the positive event.}
		By adding the constraint on the correct classification of known samples, the OSCR metric measures the trade-off between open and closed-set performance and provides an overall evaluation of the entire system.
		{For an overall characterization of the open-set classification performance of the system, we also report the TPR and CCR values obtained by fixing the threshold  $\nu$ based on the FPR.}

		\section{Experimental Results}
		\label{sec:ganresults}

		\begin{table*}[!t]
			\renewcommand\arraystretch{1.75}
			\caption{Performance of architecture attribution in closed-set (Accuracy) and open-set (AU-ROC, AU-OSCR) settings. The best results are shown in bold (the second-best is underlined). {The average EER of out-of-set detection is also reported.}
				\label{tab:table3}}
			\centering
			\large
			\resizebox{0.95\linewidth}{!}{
				\begin{tabular}{l|>{\columncolor{gray1}}c|>{\columncolor{green}}c|>{\columncolor{LightCyan}}c|>{\columncolor{gray1}}c|>{\columncolor{green}}c|>{\columncolor{LightCyan}}c|>{\columncolor{gray1}}c|>{\columncolor{green}}c|>{\columncolor{LightCyan}}c|c|c|c|c}
					\hline
					\multirow{3}{*}{Methods} & \multicolumn{3}{>{\columncolor{white}}c|}{Config S1} & \multicolumn{3}{>{\columncolor{white}}c|}{Config S2} & \multicolumn{3}{>{\columncolor{white}}c|}{Config S3} & \multicolumn{3}{c}{Average}\\
					\cline{2-14}
					& \multicolumn{1}{>{\columncolor{white}}c|}{Closed-set}& \multicolumn{2}{>{\columncolor{white}}c|}{Open-set} & \multicolumn{1}{>{\columncolor{white}}c|}{Closed-set} & \multicolumn{2}{>{\columncolor{white}}c|}{Open-set} & \multicolumn{1}{>{\columncolor{white}}c|}{Closed-set} & \multicolumn{2}{>{\columncolor{white}}c|}{Open-set} & Closed-set& \multicolumn{3}{c}{Open-set}\\
					\cline{2-14}
					& Accuracy & AU-ROC & AU-OSCR & Accuracy & AU-ROC & AU-OSCR & Accuracy & AU-ROC & AU-OSCR & Accuracy & AU-ROC  & {EER}& AU-OSCR\\
					\hline
					\hline
					{ARPL} \cite{chen2021} & \textbf{100} & 79.59 & 79.55 & 99.94 & 77.84 & 77.78 & 99.98 & 83.01 & 83.00 & 99.97 & 80.15 & {33.81} & 80.11\\
					{AKPF} \cite{xia2023} & \underline{99.95} & 75.27 & 75.27 & \textbf{100} & \underline{88.98} & \underline{88.98} & 99.56 & \textbf{91.89} & \textbf{91.6} & 99.84 & \underline{85.38} & {\underline{24.76}} & \underline{85.28}\\
					{PCSSR} \cite{huang2022} & 99.47 & \underline{83.11} & \underline{82.88} & 99.57 & 68.58 & 68.50 & 98.55 & 71.32 & 70.74 & 99.20 & 74.34 & {34.11} & 74.04\\
					{RCSSR} \cite{huang2022} & 99.62 & 82.65 & 82.46 & 99.21 & 57.98 & 57.84 & 98.65 & 70.79& 70.32 & 99.16 & 70.47 & {25.69} & 70.21\\
					\hline
					Res50Vit \cite{wang2023cvpr} & 99 & 79 & 78.32 & 99 & 76 & 75.89 & 99 & 68 & 67.93 & 99 & 74.33 & {32.50} & 74.05\\
					SiaVerify \cite{lydia2023prl} & \textbf{100} & 82.24 & 82.31 & \textbf{100} & 82.44 & 82.41 & \textbf{100} & 82.98 & 82.89 & \textbf{100} & 82.55 & {27.47} & 82.54\\
					{POSE} \cite{yang2023cvpr} & 98.56 & 75.97 & 75.60 & 96.90 & 86.73 & 85.53 & 96.70 & 83.00 & 81.50 & 97.39 & 81.90 & {26.30} & 80.88\\
					\hline
					\hline
					Baseline & \textbf{100} & {82.62} & {82.56} & \textbf{100} & {73.10} & {73.10} & \underline{99.99} & 65.99 & 65.98 & \underline{99.99} & {73.90} & {33.64} & {73.88}\\
					BOSC (Ours) & \textbf{100} & \textbf{95.31} & \textbf{95.31} & \underline{99.95} & \textbf{95.43} & \textbf{95.41} & 99.96 & \underline{90.00} & \underline{89.99 }& 99.96 & \textbf{93.58} & {\textbf{14.65}} & \textbf{93.42}\\
					\hline
			\end{tabular}}
		\end{table*}
		
		\begin{figure*}[!t]
			\centering
			\includegraphics[width=0.95\linewidth]{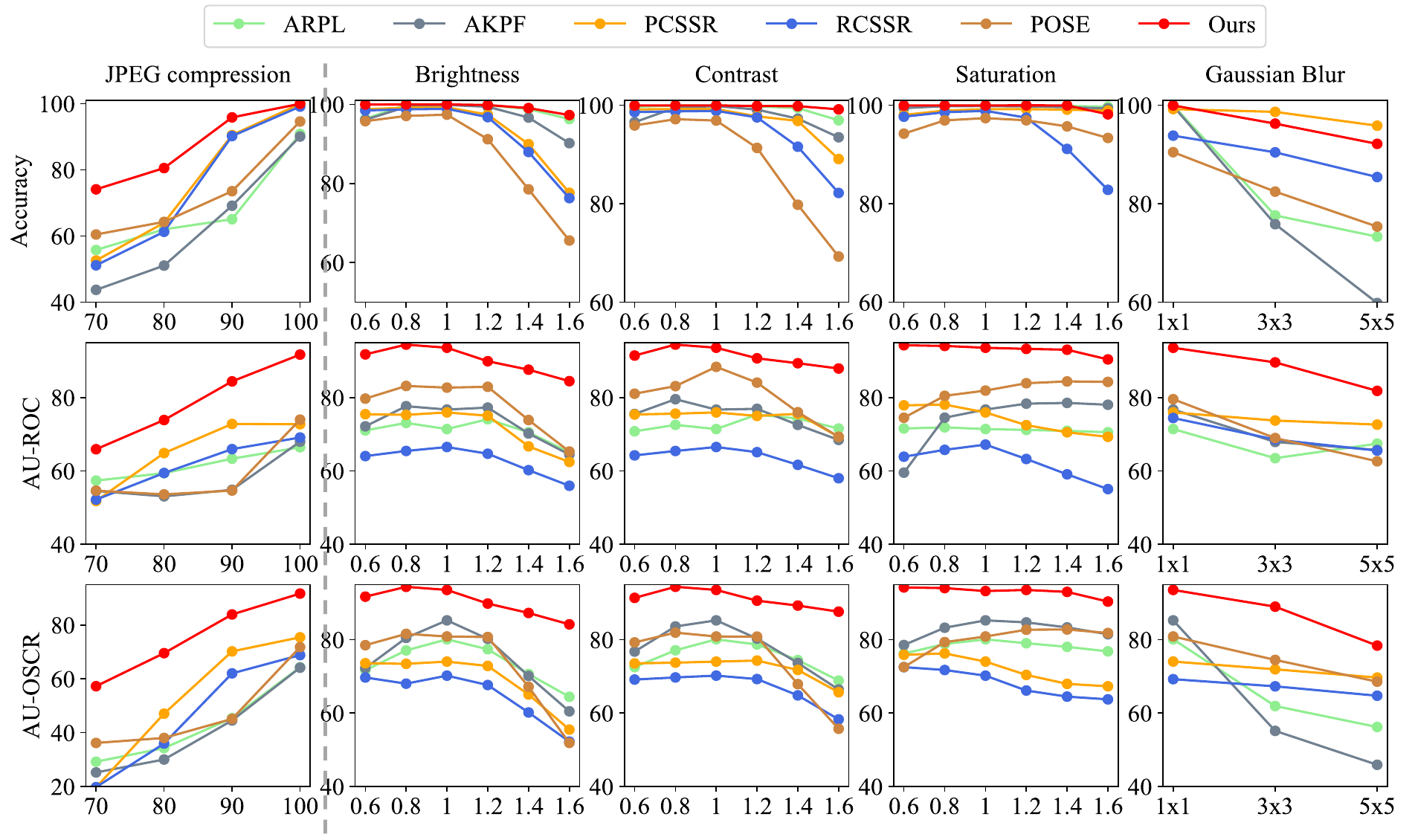}
			\caption{Robustness to image processing. From left to right: JPEG compression, brightness, contrast and saturation change, and Gaussian blur. From top to bottom: Closed-set: Accuracy, Open-set: AU-ROC and AU-OSCR. {The metric in the Y-axis on the left applies to every plot in the row. A vertical dashed line separates JPEG compression, which was considered as augmentation method during training, from the other processing, not considered during training.}}
			\label{fig:robustbcsj}
		\end{figure*}

		\begin{table}[!t]
			\renewcommand\arraystretch{1.75}
			\caption{TPR and CCR when FPR=5\% for architecture attribution in open-set. The values of the threshold $\nu$ are also reported. The best results are shown in bold (the second-best is underlined). \label{tab:newattribution}}
			\centering
			\large
			\resizebox{1\linewidth}{!}{
				\begin{tabular}{l|>{\columncolor{gray1}}c|>{\columncolor{green}}c|>{\columncolor{LightCyan}}c|>{\columncolor{gray1}}c|>{\columncolor{green}}c|>{\columncolor{LightCyan}}c|>{\columncolor{gray1}}c|>{\columncolor{green}}c|>{\columncolor{LightCyan}}c|c|c}
					\hline
					\multirow{3}{*}{Methods} & \multicolumn{3}{c|}{Config S1} & \multicolumn{3}{c|}{Config S2} & \multicolumn{3}{c|}{Config S3} & \multicolumn{2}{c}{Average}\\
					\cline{2-12}
					& \multicolumn{3}{c|}{Open-set} & \multicolumn{3}{c|}{Open-set} & \multicolumn{3}{c|}{Open-set} & \multicolumn{2}{c}{Open-set}\\
					\cline{2-12}
					& TPR & CCR &$\nu$& TPR & CCR &$\nu$ & TPR & CCR &$\nu$ & TPR & CCR\\
					\hline
					\hline
					{ARPL} \cite{chen2021} & 37.0 & 94.9 &15.2& 41.2 & 94.9 &15.0& \underline{63.1} & 94.8 &14.8 & 47.1 & 94.8\\
					{AKPF} \cite{xia2023} & 25.2 & 95.4 &-17.7& 58.4 & \textbf{95.5} &9.4& \textbf{68.5} & 95.2 &11.5& 50.7 & 95.4\\
					{PCSSR} \cite{huang2022} & 31.7 & 91.3  &-5.2& 17.1 & 92.0 &1.5& 12.9 & 86.9 &2.7& 20.6 & 90.1\\
					{RCSSR} \cite{huang2022} & 22.2 & 90.7 &-5.1& 7.9 & 92.3 &-5.7& 12.4 & 87.9 &-3.6& 14.1 & 90.3\\
					\hline
					Res50Vit \cite{wang2023cvpr} & 38.5 & 94.4 &8.9& 42.6 & 94.6 &10.5& 54.2 & 94.7 &8.9& 45.1 & 94.6\\
					\hline
					SiaVerify \cite{lydia2023prl} & 45.4 & 94.4 &0.5& 45.3 & 94.8 &0.5& 45.4 & 94.4 &0.5& 45.3 & 94.5\\
					\hline
					POSE \cite{yang2023cvpr} & 32.3 & 94.6 &0.5& 42.3 & 93.5 &0.4& 49.8 & 93.5 &0.5& 41.5 & 93.9\\
					\hline
					\hline
					Baseline & \underline{65.4} & \textbf{95.5} &11.8& 35.5 & \textbf{95.5} &7.7& 56.7 & \underline{95.5} &11.7& \underline{52.5} & \textbf{95.5}\\
					BOSC (Ours) & \textbf{79.6} & \textbf{95.5} &13.2& \textbf{80.2} & \textbf{95.5} &11.3& 59.6 & \textbf{95.5} &12.0& \textbf{73.2} & \textbf{95.5}\\
					\hline
			\end{tabular}}
		\end{table}

		\subsection{Performance analysis}
		\label{sec:gancomp}
		
		The closed-set and open-set performance of the proposed method in all three configurations is reported in Table \ref{tab:table3}, together with the performance of the state-of-the-art methods described in Section \ref{sec:sota}. We see that all methods achieve nearly perfect accuracy in the closed-set scenario, while the open-set performance are different. In particular, CSSR methods show limited effectiveness, as well as ResVit50. {The best open-set results are achieved with AKPF, SiaVerify and POSE, with the last two being specifically developed for OSIA. Notably, the} best performing state-of-the-art method is AKPF, which achieves an average AU-ROC equal to 85.38\% and an AU-OSCR equal to 85.28\%. However, this method 
		exhibit very unstable open-set performance  across the three configurations, e.g. for {AKPF}, the AU-ROC ranges from 75.27\% in Config S1 to 91.89\% in Config S3, while results tend to be more stable for SiaVerify and POSE. A possible reason is that, being based on the synthesis of open-set samples, {AKPF} is more sensitive to the composition of the unknown class and performance are very good when in-set and out-of-set classes are very different. In Config S3, where AKPF got the best performance, only GAN architectures  are included in the in-set, while LD and transformers architectures are all in the out-of-set, thus corresponding to an easy configuration
		Regarding BOSC, it achieves the best open-set performance (AU-ROC = 93.58\% and AU-OSCR= 93.42\% on average), with very limited variability across the configurations. In particular, BOSC outperforms the best-performing state-of-the-art method AKPF with a gain of 8.20\% and 8.14\% in AU-ROC and AU-OSCR, respectively. The baseline is also significantly outperformed, with a gain of approximately 20\% in both AU-ROC and AU-OSCR, confirming the effectiveness of the backdoor-based framework we have introduced.
		The table also reports the average Equal Error Rate (EER) of the ROC curve in the second-last column, which in our case is 10\% lower than AKPF (second best method). The TPR and CCR achieved by the various methods when the threshold $\nu$ is fixed to achieve the target FPR= 5\% are reported in Table \ref{tab:newattribution}, along with the corresponding values of $\nu$. We see that BOSC achieves both the highest TPR and CCR. The TPR in particular is more than 20\% larger than for the other methods.

		Fig. \ref{fig:conf-matrix} reports the confusion matrices of open-set classification for the BOSC method for all the configurations, showing the distribution of the errors among the classes, when FPR=5\%. The diagonal (1-5) reports the CCR  for every in-set class, while the last diagonal element shows the classification accuracy (accuracy of rejection) of the samples from out-of-set, i.e. the TPR. We observe that, for samples from in-set classes, the classification errors are always in favor of the rejection class, i.e., all the samples which are accepted as in-set are then correctly classified.

		Overall, the results we got show that BOSC
		significantly improves the performance of OSIA. Although it is difficult to understand exactly where the advantage of BOSC comes from, the main reasons for its good performance can be identified as follows. With respect to methods based on output score  thresholding (like  MSP and MLS thresholding), the creation of a link between in-set classes and triggers allows to differentiate better between samples from out-of-set classes and hard-to-classify samples of in-set classes, to which the closed-set model assigns a low confidence. Another advantage comes from the reduction of the open space risk. The necessity of recognizing the presence of the triggers, whose content departs significantly from the content of the query images, forces the classifier to enlarge the latent space, hence increasing its capacity to distinguish between samples form in-set and out-of-set classes. Eventually, we argue that the necessity to create a distinct link between the samples of different classes and the various triggers, forces the classifier to compact the representation of in-set classes in the latent space, hence contributing to further reduce the open space risk.
		
		\begin{figure}[!t]
			\centering
			\includegraphics[width=0.9\linewidth]{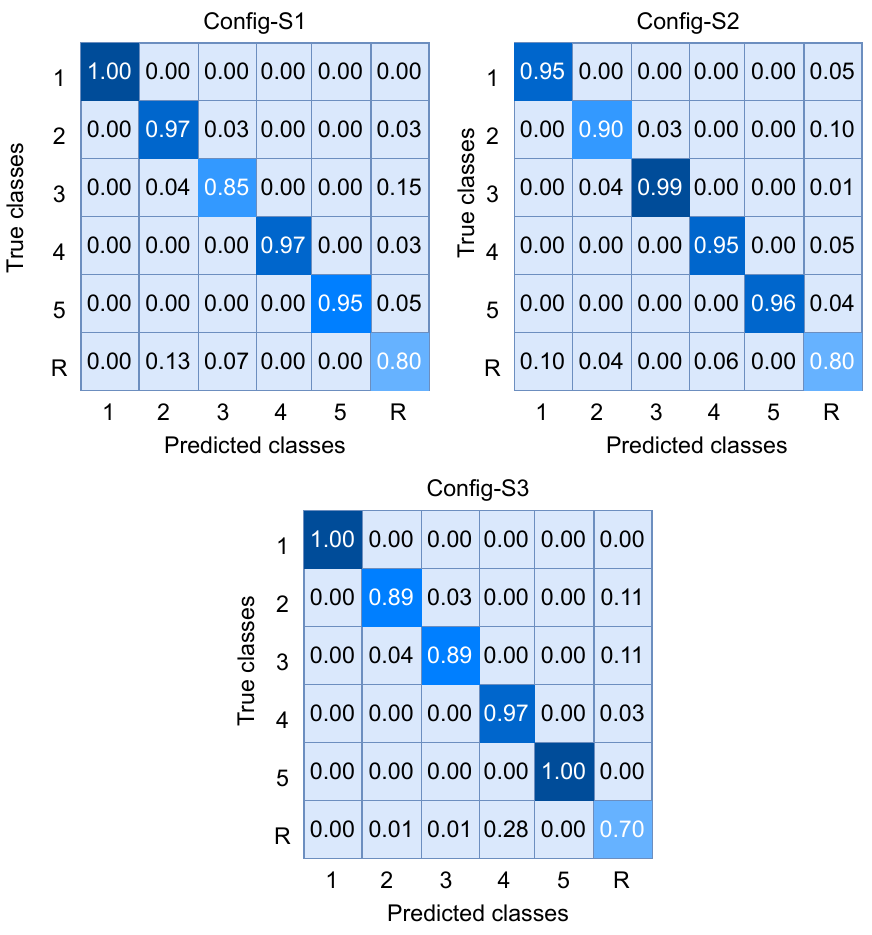}
			\caption{Confusion matrices of the BOSC method for open-set classification, when FPR is fixed to 5\%.}
			\label{fig:conf-matrix}
		\end{figure}
		
		\subsection{Robustness to image processing manipulations}
		\label{sec:ganrobust}
		
		\begin{figure}[!t]
			\centering
			\includegraphics[width=0.9\linewidth]{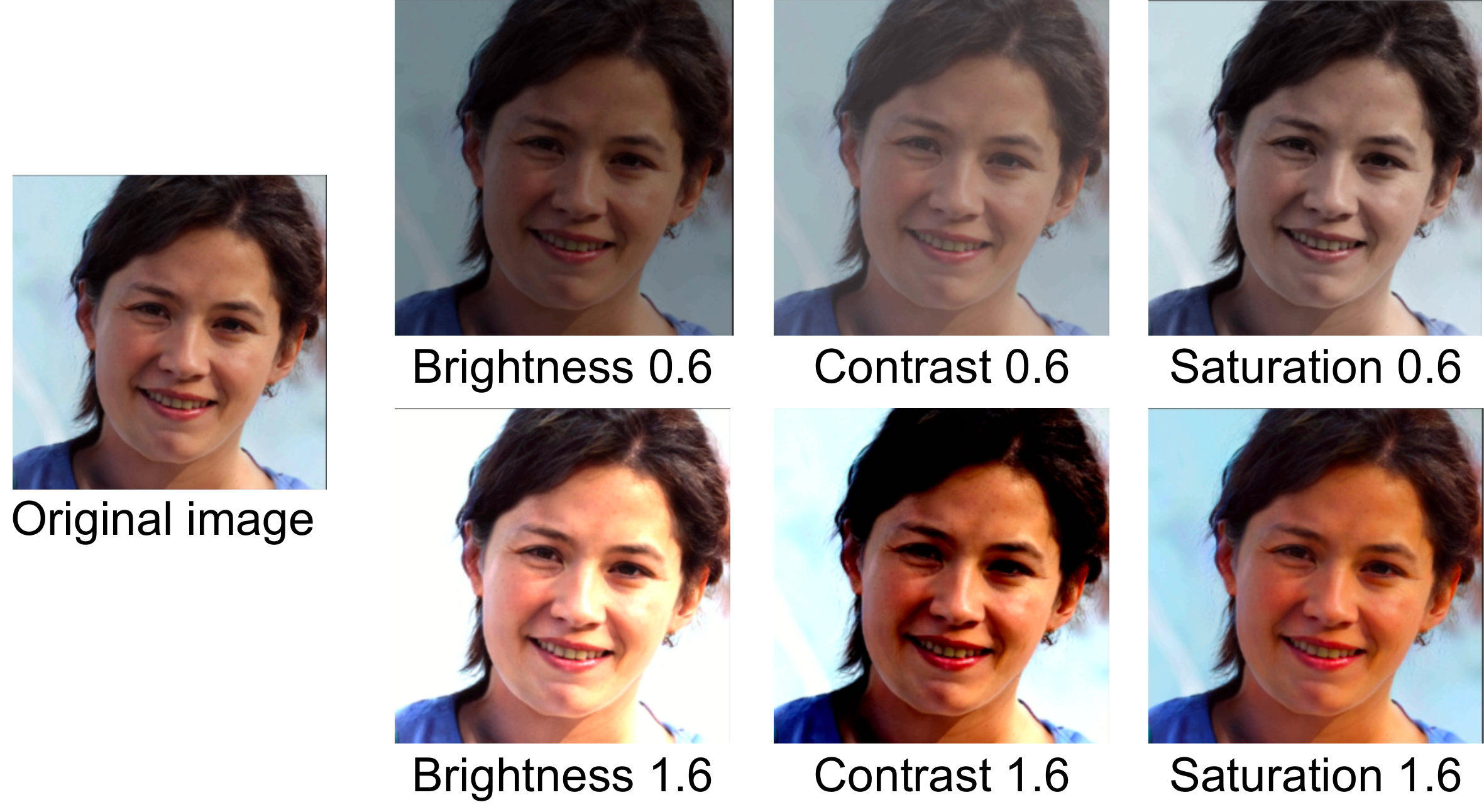}
			\caption{Examples of images processed with brightness, contrast, and saturation changes.}
			\label{fig:attack}
		\end{figure}
		In this section, we evaluate the robustness in the presence of image processing operators applied to synthetic images. In particular, we consider color modifications (saturation, brightness, contrast), Gaussian blur and JPEG compression.
		We point out that only JPEG compression has been considered during the training of our method (see Section \ref{sec:implement}), while the others correspond to never-seen processing operations.
		For the color modifications, an example of a processed image is reported in Fig. \ref{fig:attack} for the extreme values of the range of parameters we considered.
		
		Fig. \ref{fig:robustbcsj} reports the closed-set and open-set performance. BOSC has robustness similar to state-of-the-art methods in the closed-set scenario, while it gets superior performance in the open-set case, with a gain in AU-ROC and AU-OSCR larger than 10\% in all the cases. This confirms the intuition that, since the trigger images are not affected by the processing (being superimposed at test time), their features are affected to a lesser extent by processing. The most critical case is JPEG compression, notwithstanding its inclusion in the training set.

		\subsection{Generalization analysis}
		\label{sec:generalization}
		
		We also evaluated the capability of BOSC of correctly attributing to the source architecture images generated by \textit{unknown} models, that is, models different than those considered during training yet sharing the same in-set architectures
		In particular, the generative models used for these tests are obtained from the in-set architectures considering: i) different training strategies (for StyleGAN2, LSGM and DDPM),
		ii) different training datasets of real images (for Latent diffusion and Taming transformer), and iii) different image resolutions (for StyleGAN3). The details of the mismatch between the generative models considered to get the images used for training and testing are provided in Table \ref{tab:crossmodel} (column 2, 3, and 4).	With regard to the models obtained considering a mismatch in the training strategy, StyleGAN-ada refers to training with adaptive augmentation for the discriminator
		\cite{karras2020training}, while DDPM-ema is obtained by training with the exponential moving average strategy \cite{busbridge2024scale}. Finally, for LSGM, the number of stages for the training is changed, from the default value 2 to 3. In the third stage, re-training is performed by training only the SGM prior, leaving the Nouveau VAE (NVAE) component fixed\footnote{See \url{https://github.com/NVlabs/LSGM} for the details.}. The models obtained with the 2-stage and 3-stage training are referred to as quantitative and qualitative models, respectively.
		
		\begin{table}[!t]
			\renewcommand\arraystretch{2.8}
			\caption{Results (closed and open-set) in the presence of a mismatch between the generative models considered during training and testing for a same architecture. Generative models considered for testing are trained considering different datasets, parameters, and training procedures w.r.t. the models considered during training (in the last line, the number in the models' names indicates the image resolution). The underlined performance refers to the average results across different configurations.\label{tab:crossmodel}
			}
			\centering
			\resizebox{\linewidth}{!}{\begin{tabular}{c|c|c|c|c|c}
					\hline
					{\bf Architecture} & \makecell{{\bf Type of} \\{\bf Mismatch}} & {\bf Train} & {\bf Test} & {\bf ACC} & {\bf AU-OSCR}\\
					\hline
					\makecell{DDPM\\(Config-S1)} & \makecell{Training \\Methodology} & DDPM & DDPM-{ema} & 100 & 85.89\\
					\hline
					\makecell{StyleGAN2\\(Config-S1\&S2\&S3)} & \makecell{Training \\Methodology} & StyleGAN2-f & StyleGAN-ada & \underline{99.84} & \underline{92.15}\\
					\hline
					\makecell{Taming Transformer\\(Config-S1)} & Real Dataset & CelebA & FFHQ & 98.20 & 82.53\\
					\hline
					\makecell{Latent Diffusion\\(Config-S1\&S2)} & Real Dataset & CelebA & FFHQ & \underline{77.63} & \underline{71.3}\\
					\hline
					\makecell{LSGM\\(Config-S2)} & \makecell{Training \\Methodology} & \makecell{Quantitative \\(2-stages)} & \makecell{Qualitative \\(3-stages)} & 99.60 & 76.46\\
					\hline
					\makecell{StyleGAN3\\(Config-S3)} & Image Resolution & \makecell{StyleGAN3\\t-1024\&\\t-ffhqu1024\&r} & \makecell{StyleGAN3\\ t-ffhqu256} & 99.35 & 78.17\\
					\hline
			\end{tabular}}
		\end{table}
		The last two columns of Table \ref{tab:crossmodel} report the closed-set accuracy and the AU-OSCR obtained for the specific in-set architecture (detailed in column 1), considering  test images obtained from mismatched generative models.
		The results show that the closed-set performance is very good (Acc above 98\%) in all cases, with the exception of Latent Diffusion, where we get Acc = 77.63\%. This confirms that the classifier indeed looks at architecture-level features, and not at some model specific features. The model mismatch affects the open-set performance more, and in fact, the  AU-OSCR computed on the mismatched samples decreases. Performance is great in the case of StyleGAN2 and remains pretty good also for DDPM and Taming Transformer, while they drop in the case of Latent Diffusion, LSGM and StyleGAN3.
		A strategy that we expect can help mitigating this issue is to include multiple models for every architecture inside the training set. Arguably, doing so should induce the system to better learn the model variability for a given generative architecture.
		
		Finally, we also ran an experiment to measure the performance in a realistic scenario where the OSIA methods are tested with pristine images. Specifically, in this experiment, we included images from two pristine datasets, namely the CelebA and the FFHQ (2.000 images per each), in the out-of-set class, which then comprises a mixture of pristine images and images generated by different generative architectures (2 pristine datasets and 5 generative architectures). Fig. \ref{fig:openreal}
		shows the results, averaged across the three configurations, in terms of AU-ROC, AU-OSCR, EER and TPR at FPR = 5\% (the FP threshold is set on the in-set class, hence it is the same as before). 
		We see that the inclusion of pristine images leads to a performance drop for many methods, especially the best performing ones. Notably, BOSC remains the method achieving the best results according to all metrics. 
		
		\begin{figure*}[h]
			\centering
			\includegraphics[width=0.47\linewidth]{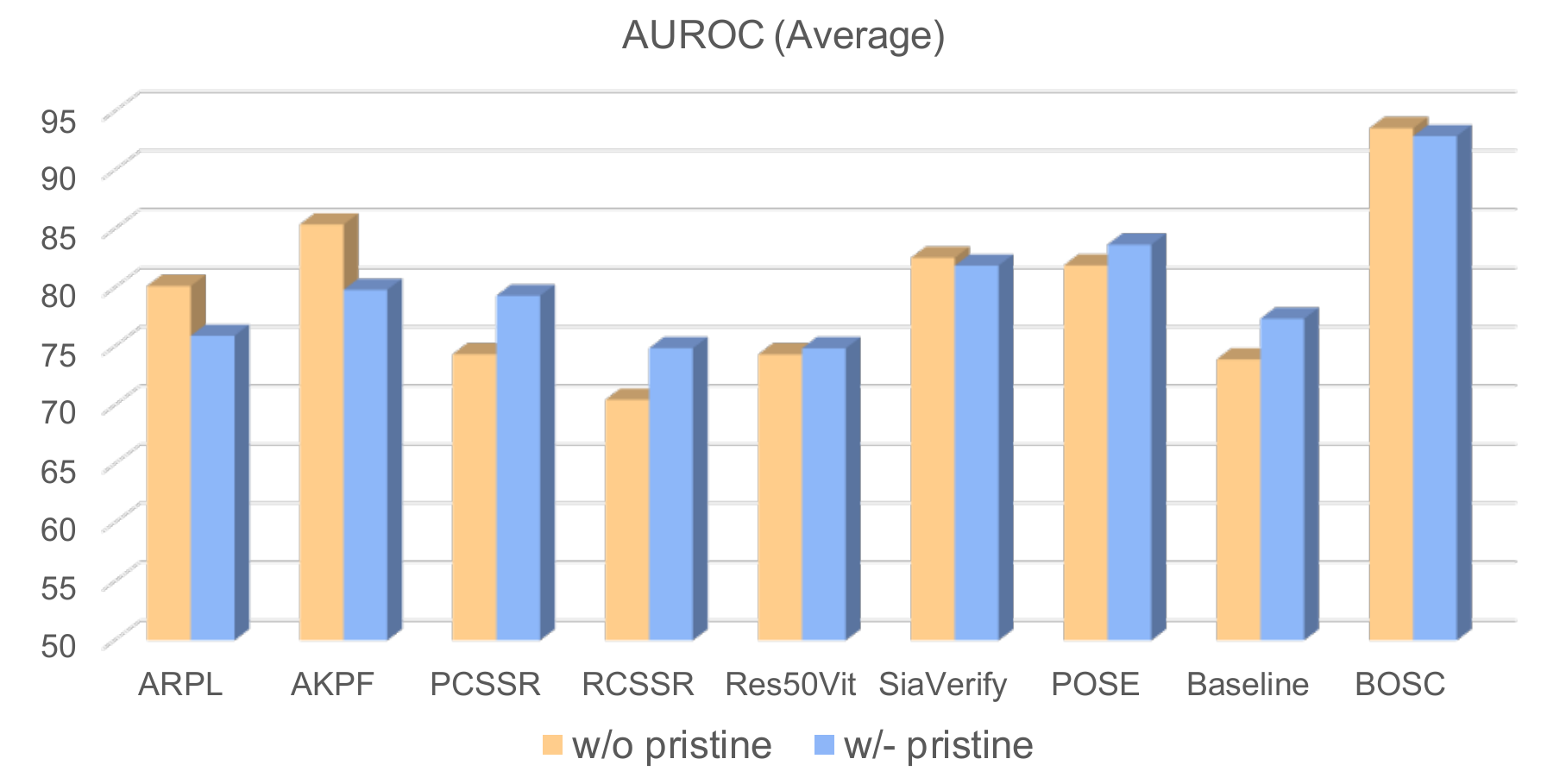}
			\includegraphics[width=0.47\linewidth]{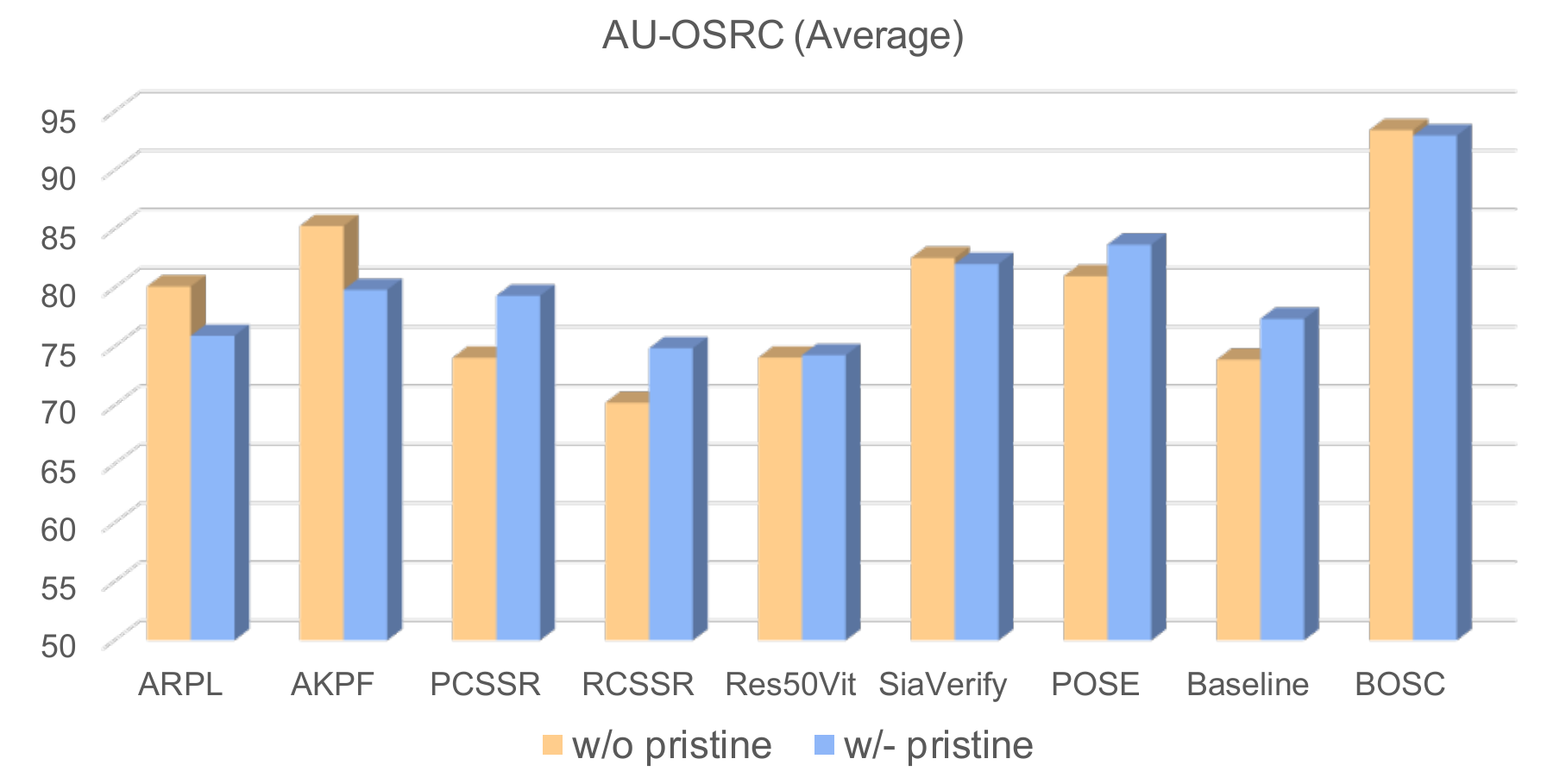}
			\includegraphics[width=0.47\linewidth]{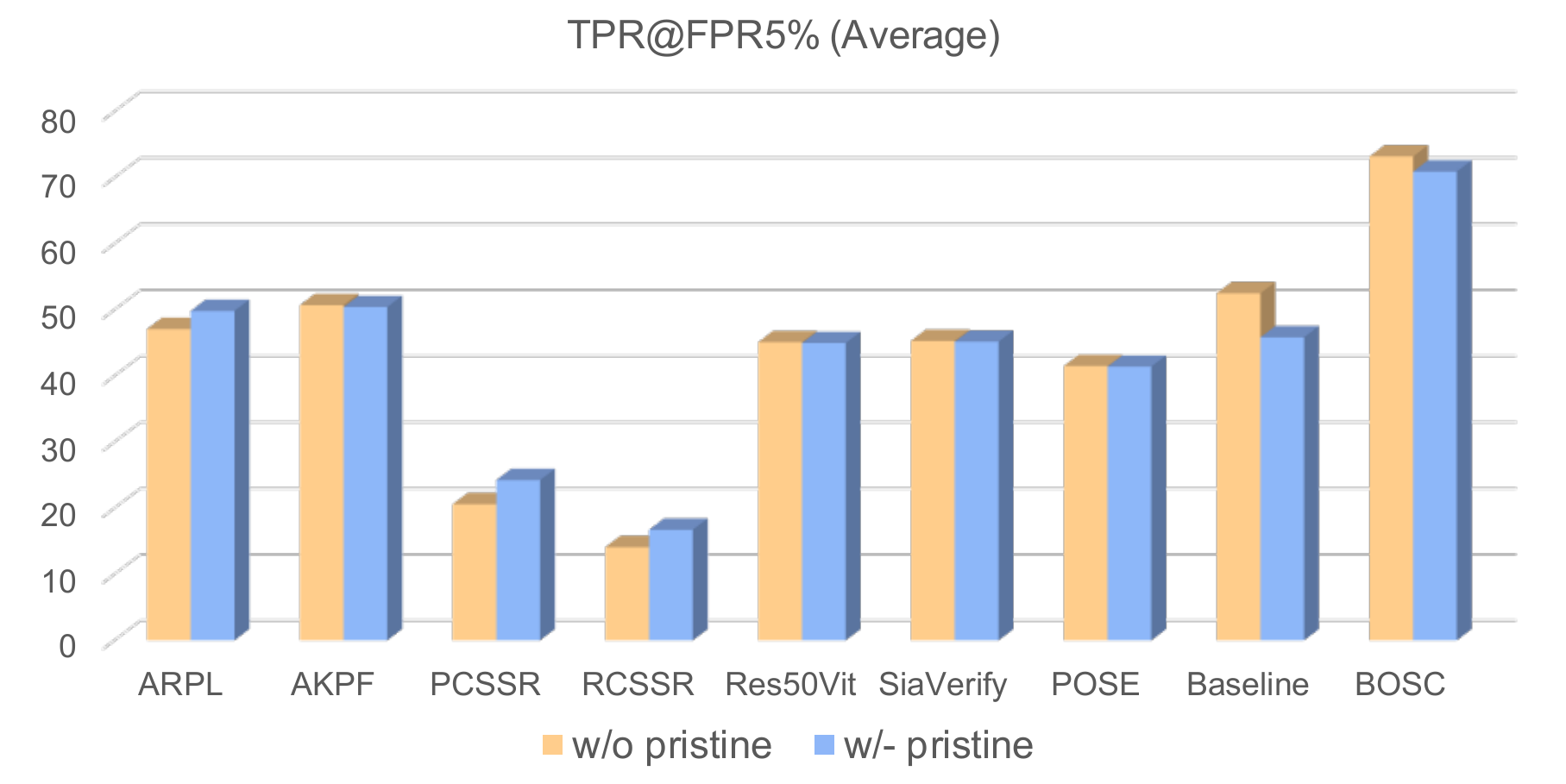}
			\includegraphics[width=0.47\linewidth]{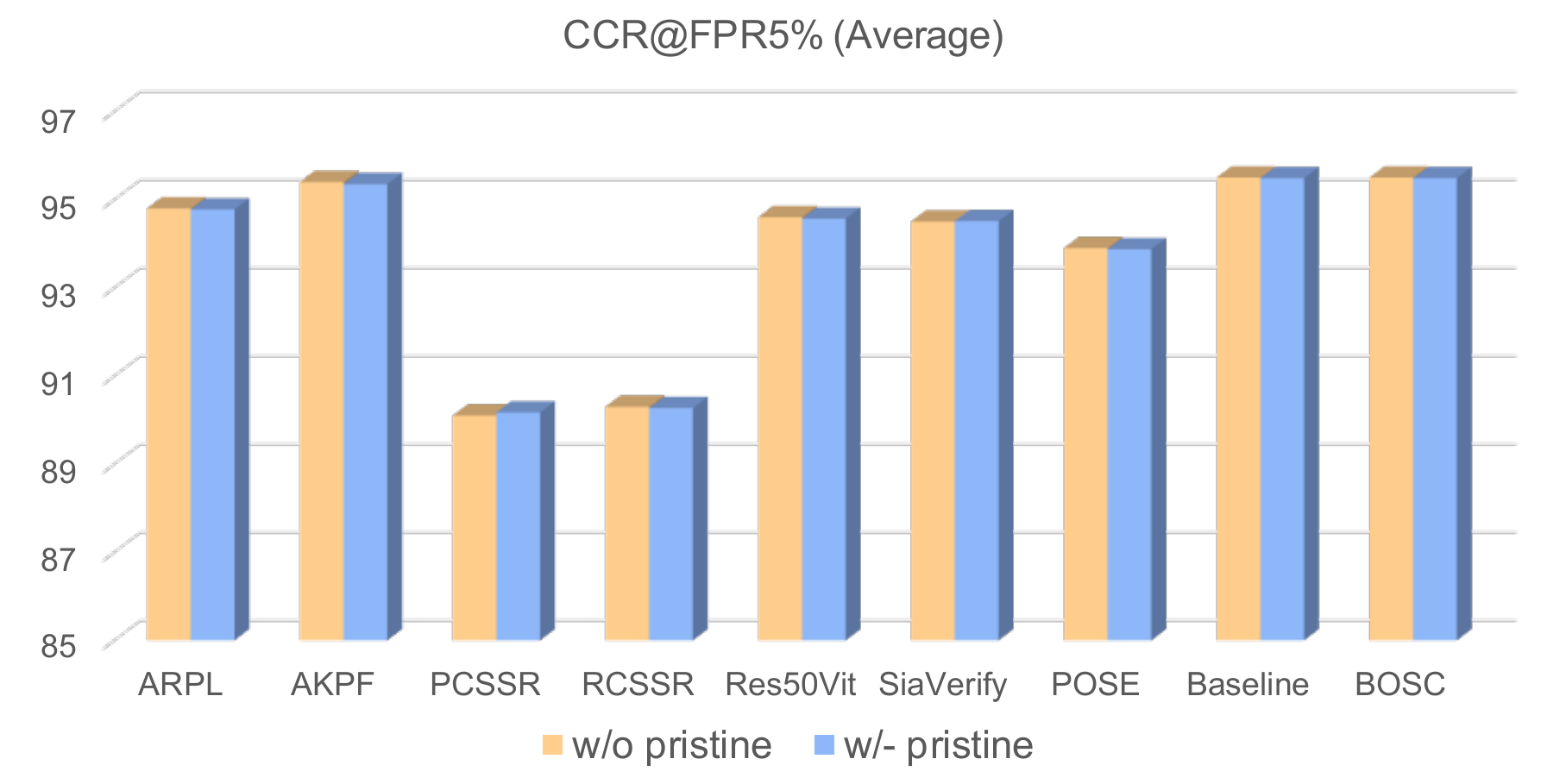}
			\caption{Average open-set performance (in terms of AU-ROC, AU-OSCR, TPR@FPR5\% and EER) of the various methods when pristine images (from CelebaHQ and FFHQ) are also included in the out-of-set class. The results without including pristine images are also reported.}
			\label{fig:openreal}
		\end{figure*}

		\subsection{Ablation study}
		\label{sec:ablationstudy}
		
		We carried out an ablation study to assess the impact of each component of BOSC.
		
		\subsubsection{Choice of rejection score}
		\label{sec:scorecomparison}
		\begin{figure}[!t]
			\centering
			\includegraphics[width=0.85\linewidth]{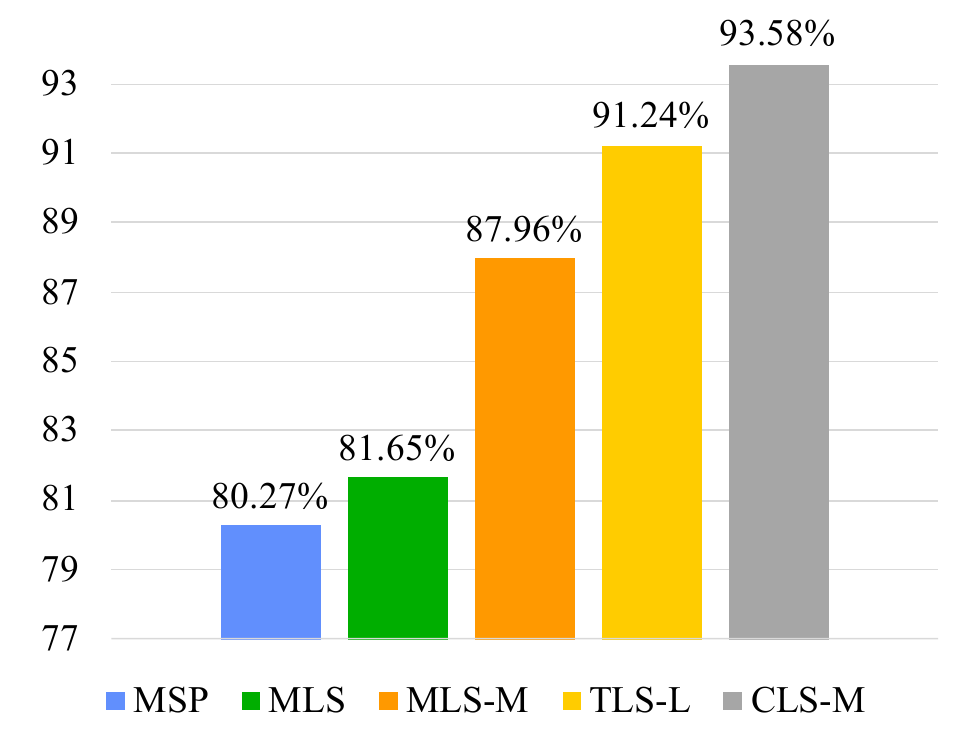}
			\caption{Average performance (AU-ROC) of out-of-set rejection of BOSC using different scores. The average is taken over the three in-set and out-of-set configurations.}
			\label{fig:scores}
		\end{figure}
		
		The benefit of considering the trigger-based score in Eq. \eqref{eq:pcls} for sample rejection with the backdoor-based network is shown in
		Fig. \ref{fig:scores}. In particular, the combined logit score (CLS-M) is compared with other trigger-based scores, that is, TLS-M and MLS-S (see section \ref{sec:OSscore} for the definition), and also scores commonly used for OSR, which are directly obtained from the prediction output vector of $x$. In particular, we considered the maximum value of the softmax probability vector (MSP), and the maximum logit score (MLS).
		
		\begin{table}[]
			\renewcommand\arraystretch{1.25}
			\centering
			\caption{Ablation study on the effect of mixup augmentation.}
			\begin{tabular}{c|c|c|c}
				\hline
				& Accuracy & AU-ROC & AU-OSCR\\
				\hline
				Baseline & 100 & 73.90 & 73.88\\
				\hline
				BOSC (w/o Mixup) & 99.98 & 86.00 & 85.99\\
				\hline
				BOSC (w/- Mixup) & 99.97 & 93.58 & 93.42\\
				\hline
			\end{tabular}
			\label{tab:abla}
		\end{table}
		
		\begin{figure}[!t]
			\centering
			\includegraphics[width=0.9\linewidth]{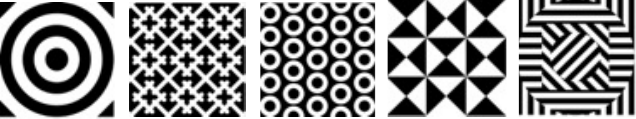}
			\caption{A different set of trigger images used in the ablation study for the attribution task.}
			\label{fig:pattern}
		\end{figure}

		We see that the AU-ROC obtained with trigger-based scores (MLS-M, TLS-M, CLS-M) is much higher than the AU-ROC obtained with common scores computed on the output of clean samples (MSP and MLS). Among the three trigger-based scores, we see that  TLS-M improves the performance of MLS-M from 87.96\% to 91.24\% on average. The performance is further improved with CLS-M (AU-ROC = 93.58\%), which fully exploits the different behaviour in the presence of matched and mismatched triggers.
		
		\begin{table}
			\renewcommand\arraystretch{2.25}
			\caption{Summary of the 19 editing classes (18 + ’None’) and the configurations of in-set (11 classes) and out-of-set (8 classes).}
			\begin{center}
				\resizebox{\linewidth}{!}{
					\begin{tabular}{c|c}
						\hline
						Editing tools & Edit types \\
						\hline
						PTI & T0: None (Reconstructed)\\
						\hline
						InterfaceGAN & T1-T4: Smile, Not smile, Old, Young \\
						\hline
						StyleCLIP&\makecell{\textbf{Expression} (T5, T6): Angry, Surprised \\ \textbf{Hair style} (T7-T12): Afro, Purple\_hair, \\ Curly\_hair, Mohawk, Bobcut, Bowlcut \\ \textbf{Identity change} (T13-T18): Taylor\_swift, Beyonce, \\ Hilary\_clinton, Trump, Zuckerberg, Depp}\\
						\hline
						\hline
						Configurations & In-set \& Out-of-set\\
						\hline
						Config F1 & \makecell{In-set: T4, T1, T2, T5, T6, T7, T8, T9, T10, T11, T12\\Out-of-set: T0, T3, T13, T14, T15, T16, T17, T18}\\
						\hline
						Config F2 & \makecell{T5, T1, T2, T3, T4, T11, T12, T13, T14, T15, T18\\Out-of-set: T0, T6, T7, T8, T9, T10, T16, T17}\\
						\hline
						Config F3 & \makecell{T2, T1, T3, T4, T6, T10, T12, T15, T16, T17, T18\\Out-of-set: T0, T5, T7, T8, T9, T11, T13, T14}\\
						\hline
				\end{tabular}}
				\label{tab:facedata}
			\end{center}
		\end{table}
		
		\subsubsection{Mixup augmentation}
		We also ran some experiments to assess the benefit of the mixup augmentation strategy. Table \ref{tab:abla} reports the closed-set and open-set performance achieved by BOSC when training is carried out with and without mixup augmentation. The performance of the baseline is also reported.  We see that, while the accuracy values of all the models are the same,  the open-set performance improve significantly with the adoption of mixup augmentation.
		In particular, the gain brought by the mixup strategy in the performance of BOSC is 7.58\% in both AU-ROC and AU-OSCR.

		\subsubsection{Choice of Triggers}%
		Finally, we run additional experiments with different trigger images, namely, using geometrical trigger patterns (shown in Fig. \ref{fig:pattern}).
		We got the following average performance: Accuracy 99.62\%, AU-ROC = 77.47\%, AU-OSCR = 77.30\%. It is worth noticing that, although this set of trigger images do not perform so well as the cartoon images (see Table \ref{tab:abla}), with a reduction of approx 15\% in both AU-ROC and AU-OSCR),  the open-set performance remains superior to those achieved by the baseline, which is surpassed by 3/4\% in both AU-ROC and AU-OSCR.
		
		\section{Application to facial editing classification}
		\label{sec:classification}
		
		In this section, we describe the experiments we ran and the results we got on an different image forensic task, namely the open-set classification of facial attribute editing \cite{wang2023icassp}, to validate the generality of the BOSC framework.
		The classification of facial attribute editing consists in detecting the particular facial attribute modified by means of a conditional generative network.  Manipulated facial attributes include age, gender, expression or hair color.
		Classifying the manipulated attributes in facial images enables effective analysis and detection of tampered or manipulated content, ensuring the integrity and authenticity of visual data in digital environments.
		
		\subsection{GAN Face Editing Dataset and Setting}
		\label{sec:facialdata}

		The dataset for these experiments is built as in \cite{wang2023cvpr}.
		The face images are taken from the CelebA-HQ dataset.
		The images are manipulated with
		18 editing types:
		4 facial attributes are edited with InterfaceGAN \cite{shenInterfacegan}, and 14 facial attributes with StyleCLIP \cite{patashnikStyleCLIP}, namely, methods for image manipulation based on StyleGAN that exploit the disentanglement of the features in the latent space, utilizing it to edit the semantic attributes through linear subspace projection.
		To get the desired editing, the Pivotal Tuning Inversion (PTI) method is considered to extract the latent code of the real images \cite{roich2022ptiI}. Then, the image attribute is manipulated via  InterfaceGAN or StyleCLIP using the learned latent code and a target StyleGAN2 generator to achieve the desired modification of the attribute.
		An overview of the dataset is provided in Table \ref{tab:facedata}. The reconstructed version of the image with no editing (`None') by PTI \cite{roich2022ptiI} is also considered, for a total of 19 classes.
		The editing attributes are grouped into	four categories: expression, aging, hairstyle and identity change.
		The editing types are split into 11 in-set and 8 out-of-set classes. Three different splittings of in-set and out-of-set editing types are considered, named `Config-F1', `Config-F2' and `Config-F3'.
		Fig. \ref{fig:faceexa} shows an example of a manipulated face image for each edited attribute.
		A total number of 5992 images were taken from CelebA-HQ dataset.
		For each configuration,
		4400 CelebA-HQ images have been used to generate the edited images used for training, 1592 for those used for testing, for a total of 83600 (4400 × 11) images for training and 30248 (1592 × 19) for testing.
		
		The BOSC classifier was trained as described in the previous section, using the same parameters' setting reported in Section \ref{sec:implement}. The
		11 trigger images used for training are illustrated in Fig. \ref{fig:triggers}. Regarding augmentation, we considered only horizontal flipping.
		
		
		\begin{figure}[!t]
			\centering
			\includegraphics[width=0.95\linewidth]{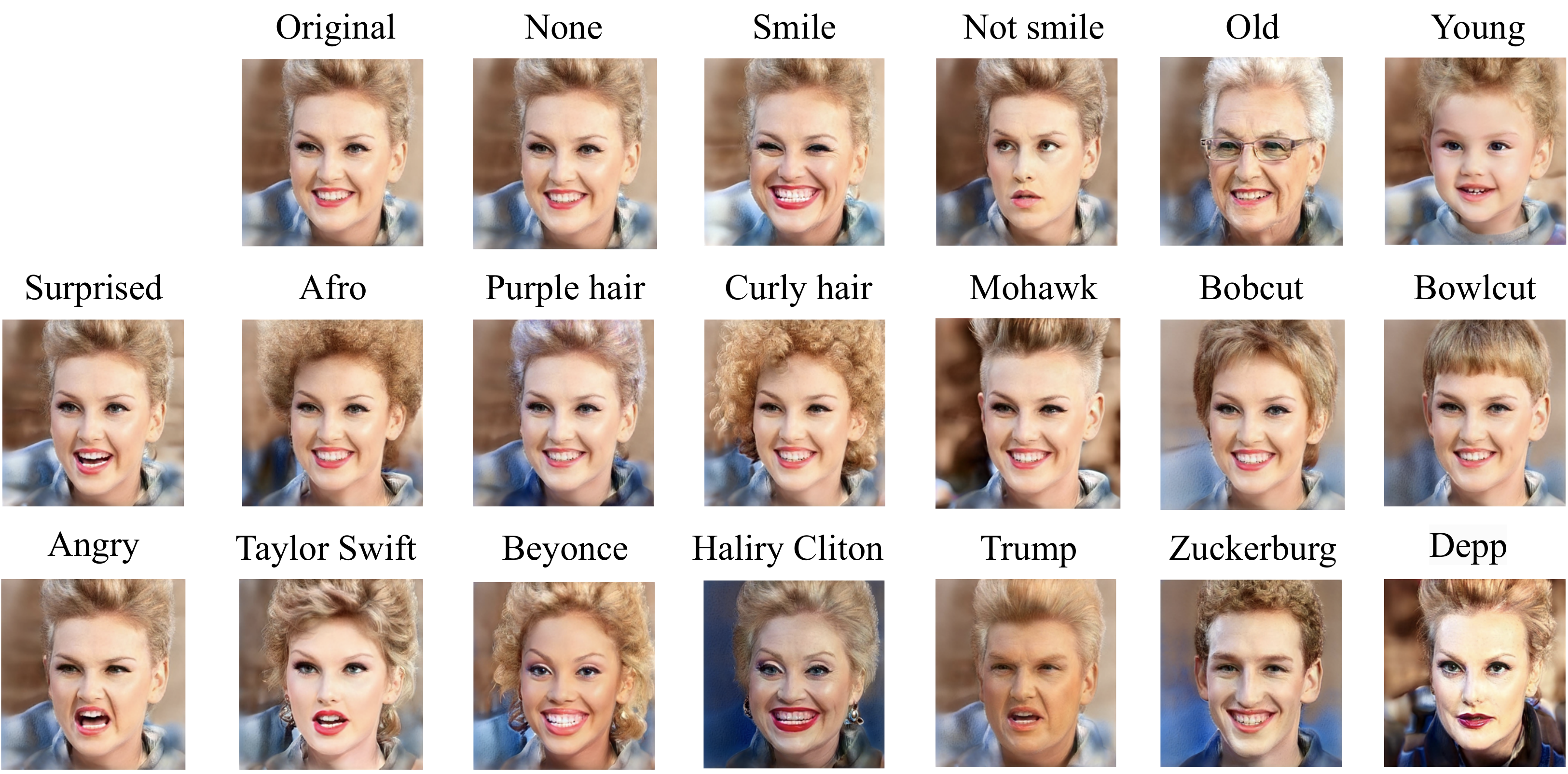}
			\caption{Examples of edited images by InterfaceGAN \cite{shenInterfacegan} (first
				row) and StyleCLIP \cite{patashnikStyleCLIP} (second and third rows).}
			\label{fig:faceexa}
		\end{figure}
		
		\subsection{Results}
		\label{sec:facialresult}
		\begin{table*}[!t]
			\renewcommand\arraystretch{1.75}
			\caption{Performance of facial editing classification in closed-set (Accuracy) and open-set (AU-ROC, AU-OSCR). The best results are shown in bold (the second-best is underlined). The average EER of out-of-set detection is also reported.
				\label{tab:editsota}}
			\centering
			\large
			\resizebox{0.95\linewidth}{!}{
				\begin{tabular}{l|>{\columncolor{gray1}}c|>{\columncolor{green}}c|>{\columncolor{LightCyan}}c|>{\columncolor{gray1}}c|>{\columncolor{green}}c|>{\columncolor{LightCyan}}c|>{\columncolor{gray1}}c|>{\columncolor{green}}c|>{\columncolor{LightCyan}}c|c|c|c|c}
					\hline
					\multirow{3}{*}{Methods} & \multicolumn{3}{c|}{Config F1} & \multicolumn{3}{c|}{Config F2} & \multicolumn{3}{c|}{Config F3} & \multicolumn{3}{c}{Average}\\
					\cline{2-14}
					& \multicolumn{1}{>{\columncolor{white}}c|}{Closed-set}& \multicolumn{2}{c|}{Open-set} & \multicolumn{1}{>{\columncolor{white}}c|}{Closed-set}& \multicolumn{2}{c|}{Open-set} & \multicolumn{1}{>{\columncolor{white}}c|}{Closed-set}& \multicolumn{2}{c|}{Open-set} & Closed-set& \multicolumn{3}{c}{Open-set}\\
					\cline{2-14}
					& Accuracy & AU-ROC & AU-OSCR & Accuracy & AU-ROC & AU-OSCR & Accuracy & AU-ROC & AU-OSCR & Accuracy & AU-ROC & EER & AU-OSCR\\
					\hline
					\hline
					{ARPL} \cite{chen2021} & {91.92} & 86.84 & 82.54 & 94.41 & 87.34 & 84.57 & 90.99 & 85.84 & 80.64 & 92.44 & 86.67 & 19.20 & 82.58\\
					{AKPF} \cite{xia2023} & 94.41 & 91.09 & \underline{87.84} & 95.33 & 88.72 & 86.49 & 91.45 & \underline{87.35} & 82.58 & 93.73 & 89.05 & 18.36 & 85.64\\
					{PCSSR} \cite{huang2022} & 95.33 & 85.60 & 82.77 & 96.40 & 82.60& 80.63 & 91.87 & 86.23 & 81.39 & 94.53 & 84.81 & 21.31 & 81.60\\
					{RCSSR} \cite{huang2022} & 95.05 & 82.46 & 79.48 & 97.02 & 89.98 & 88.37 & 93.27 & 82.68 & 78.41 & 95.11 & 85.04 & 22.84 & 82.09\\
					\hline
					Res50Vit \cite{wang2023cvpr} & 93.65 & \underline{91.42} & 87.65 & 95.59 & \textbf{91.66} & \underline{89.50} & 91.83 & 86.64 & 82.13 & 93.69 & \underline{89.91} & \underline{17.83} & \underline{86.43}\\
					\hline
					\hline
					Baseline & \textbf{97.21} & 87.85 & 86.66 & \textbf{97.91} & 88.22 & 87.24 & \textbf{95.09} & 86.10 & \underline{83.92} & \textbf{96.74} & 87.39 & 18.03 & 85.94\\
					BOSC (Ours) & \underline{96.65} & \textbf{92.13} & \textbf{90.35} & 97.28 & \underline{91.62} & \textbf{90.49} & \underline{94.50} & \textbf{88.43} & \textbf{85.40}& \underline{96.16} & \textbf{90.73} & \textbf{16.49} & \textbf{88.75}\\
					\hline
			\end{tabular}}
		\end{table*}
		\begin{figure*}[!t]
			\centering
			\includegraphics[width=0.95\linewidth]{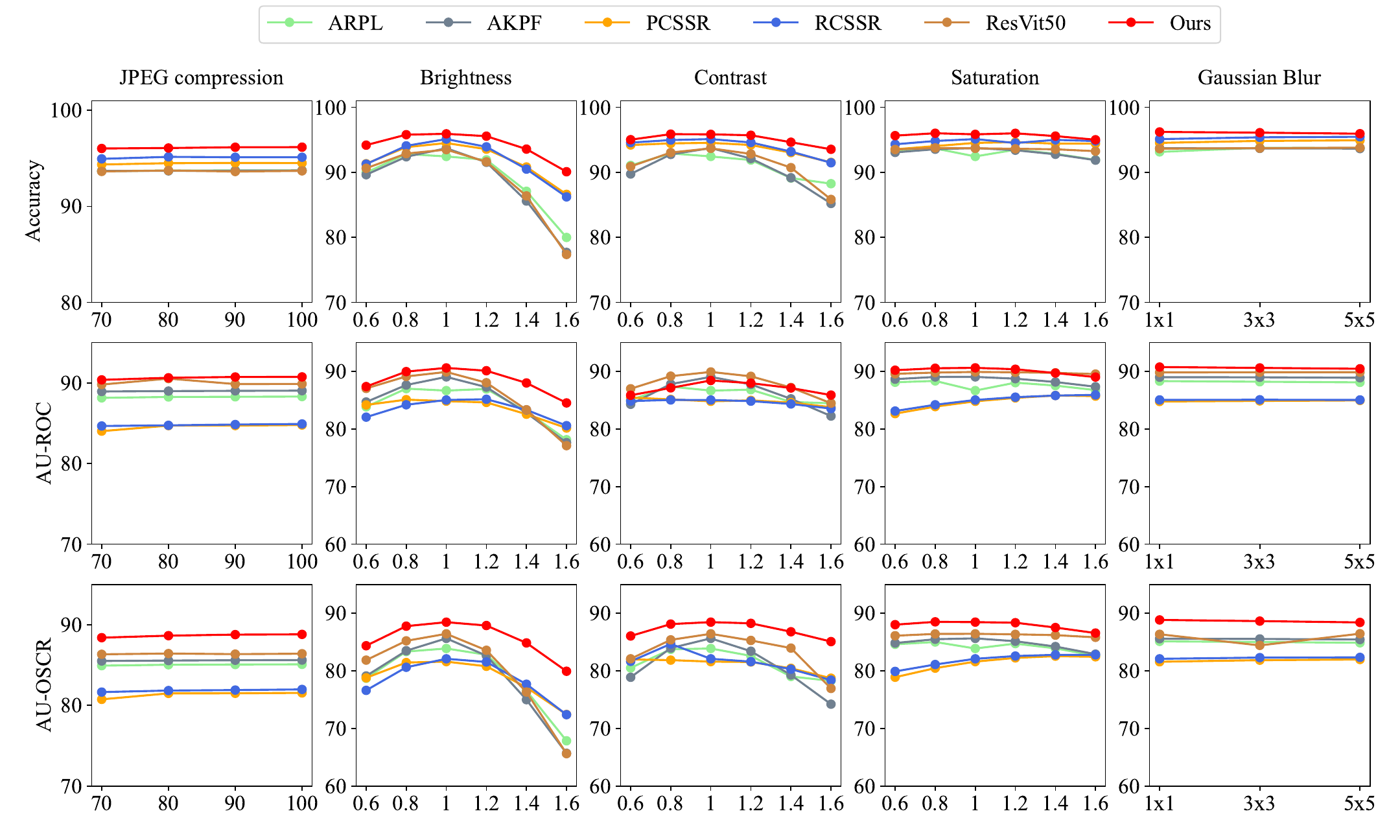}
			\caption{Robustness to image-level attacks. From left to right: JPEG compression, brightness, contrast, saturation and Gaussian blur. From top to bottom:  Accuracy (closed-set) AU-ROC and  AU-OSCR (open-set).
			}
			\label{fig:robustface}
		\end{figure*}

		The proposed method is compared with {ARPL} \cite{chen2021}, {AKPF} \cite{xia2023}, {PCSSR} \cite{huang2022}, {RCSSR} \cite{huang2022} (general OSR methods),
		and the ResVit50 method \cite{wang2023cvpr}, which, as pointed out before, corresponds to a method specifically proposed for this task.
		The performance of an EfficientNet-B4 trained on the in-set classes using MLS for out-of-set detection (baseline) is also reported.
		
		The results achieved for in-set and out-of-set classes are shown in Table \ref{tab:editsota}.
		%
		We see that all methods obtain AU-ROC and AU-OSCR larger than 80\%, while the closed-set accuracy of the various methods ranges between 92\% and 97\%, with BOSC performing the best in almost all the cases.
		Besides, the results of all methods are stable across the various configurations. The best open-set results for the state-of-the-art are achieved by {AKPF} and Res50Vit, with the latter yielding the best performance.
		Compared to Res50Vit, BOSC achieves a small gain of
		0.82\% in AU-ROC, 1.34\% in EER, and  2.32\% in AU-OSCR (and a 2.47\% gain in the Accuracy). 
		We stress that, while AKPF (as well as ARPL and the CSSR methods) is a  general algorithm for OSR,  Res50Vit is a method  designed for  open-set facial editing classification, hence specialized for this task. This method resorts to the aid of a localization branch to focus on the face regions that are most relevant for the various editing. Therefore, the similar (slightly better) performance obtained by BOSC represents a noticeable result, also proving its generality across different image forensic tasks. In addition, we applied BOSC to this task using the same set of trigger images adopted for architecture attribution. However, better performance could be achieved with a different set (the optimization of the set of trigger images is left to future work).
		
		With regard to the performance at a fixed threshold, BOSC achieves an average CCR of 93.6\% and a TPR of 52.9\% at FPR=5\%. The threshold values $\nu$ are in the range 9-11, which is similar to the range of values for the architecture attribution task.

		
		Finally, the robustness performance of the various methods against brightness, contrast, saturation, Gaussian blur and JPEG compression are reported in Fig. \ref{fig:robustface}. We see that BOSC always achieves the best robustness. In the case of contrast adjustment, {AKPF} and Res50Vit slightly outperform BOSC in terms of out-of-set detection capability (AU-ROC). However, the overall classification performance in open set (AU-OSCR) is noticeably superior for BOSC.
		

		\section{Conclusion}
		\label{sec:conclusion}
		We have presented a backdoor-based open-set classification (BOSC) framework for open-set synthetic image attribution.
		BOSC works by injecting class-specific triggers inside a
		portion of the images of the training set to induce the network to
		establish a link between class features and trigger features. Such a matching is exploited during testing for out-of-set detection by means of an ad-hoc score.
		%
		The results show that the proposed method achieves better open-set performance compared to the state-of-the-art.
		The effectiveness of BOSC was also validated on
		the open-set classification of synthetic facial editing,  confirming the superiority over the state-of-the-art also on this task, as well as the generality of the approach.
		
		Future work will focus on investigating the impact of the choice of trigger images on the results and on optimizing the triggers for a given classification task. 
		Notably, our experiments revealed that not all triggers are equally effective in the context of OSIA. Unlike malevolent backdoor attacks, where stealthiness is a  major requirement, OSIA triggers do not need to be hidden - making their visibility acceptable and even potentially beneficial. In fact, we observed that semantically meaningful triggers (e.g., cartoon images) consistently outperform low-level, texture-based triggers in rejecting unknown classes. This insight suggests that the nature of the trigger plays a crucial role in OSIA performance and opens a new research direction: the design of specific, semantically rich triggers to maximize open-set attribution performance. 
		We also plan to apply BOSC to other open-set classification tasks, also beyond image forensics, to validate its effectiveness for general OSR image classification tasks.

		%
		%
		
		%
		%
		%


		\newpage
		
		
		\begin{IEEEbiography}[{\includegraphics[width=1in,height=1.25in,clip,keepaspectratio]{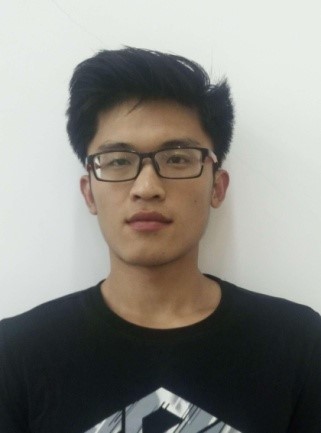}}]{Jun Wang} obtained his PhD degree from the Department of Information Engineering and Mathematics, University of Siena in 2024, M.S. degree and B.S. degree from Shandong Normal University, China, in 2017 and 2020. He is currently a Post-Doc at Great Bay University. His research interests include multimedia forensics, digital watermarking, image processing.
		\end{IEEEbiography}
		\begin{IEEEbiography}[{\includegraphics[width=1in,height=1.25in,clip,keepaspectratio]{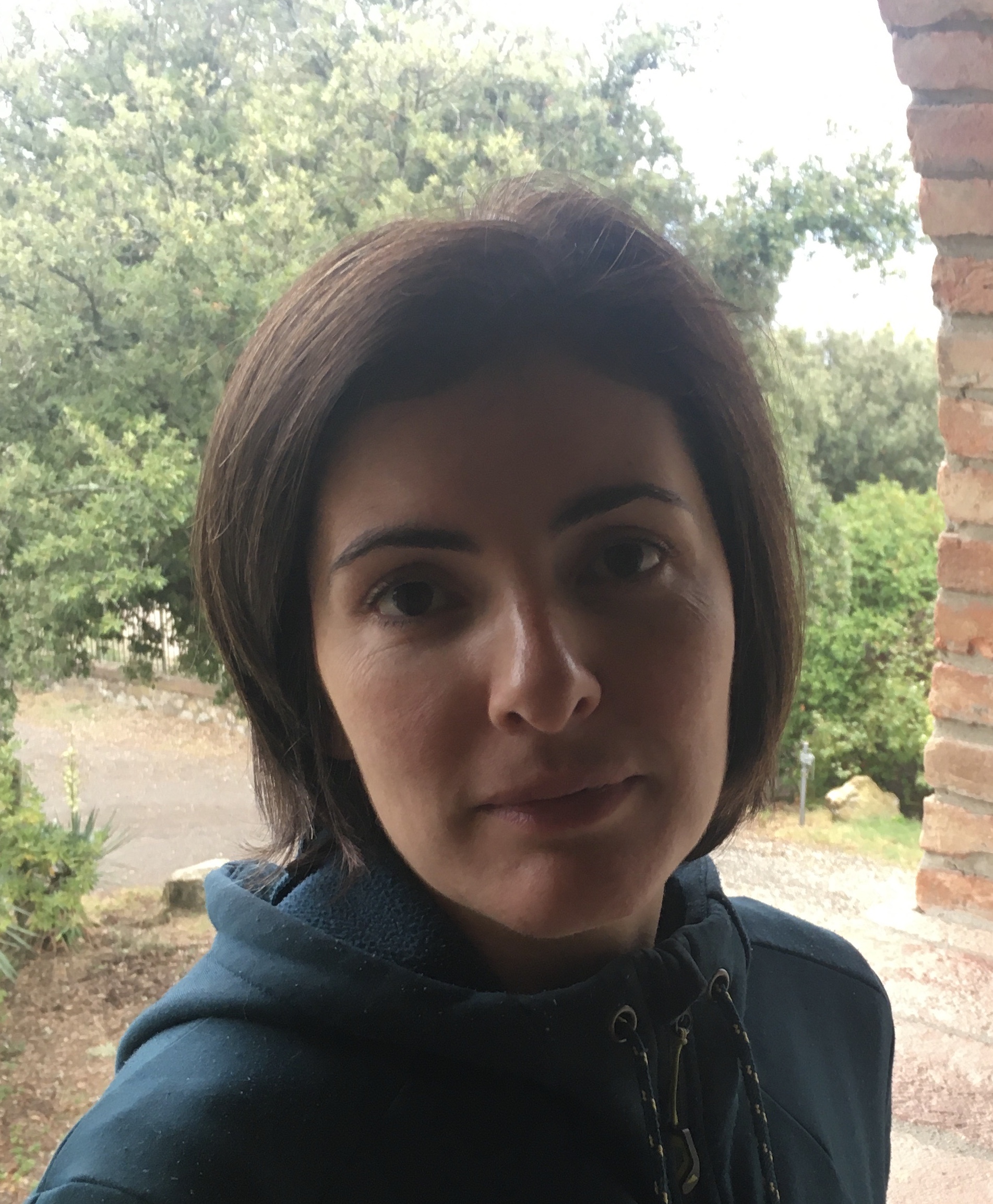}}]{Benedetta Tondi} received the MSc degree in electronics and communications engineering, and the PhD degree in information engineering and mathematical sciences, from the University of Siena, Italy, in 2012 and 2016, respectively. She is currently an Associate Professor at the Department of Information Engineering and Mathematics, University of Siena. She is a member of the Information Forensics and Security Technical Committee of the IEEE Signal Processing Society. Her research interests focus on adversarial signal processing, multimedia forensics, AI security, and deep neural networks watermarking. She currently serves as an Associate Editor for the IEEE Transactions on Information Forensics and Security and the IEEE Signal Processing Letters. She has been Technical Program Chair of ACM IH\&MMSEC 2022 and Area Chair of several IEEE conferences and workshops. She has received Best Paper Awards at WIFS and MMEDIA. She is the recipient of the 2017 GTTI PhD Award for the best PhD thesis defended at an Italian University in the areas of Communications Technologies (Signal Processing, Digital Communications, Networking).
		\end{IEEEbiography}
		
		\begin{IEEEbiography}[{\includegraphics[width=1in,height=1.25in,clip,keepaspectratio]{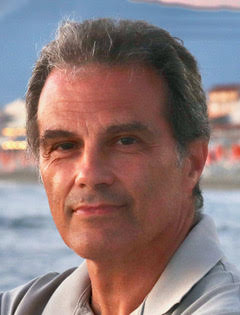}}]{Mauro Barni} graduated in electronic engineering at the University of Florence in 1991. He received the PhD in informatics and telecommunications in October 1995. During the last two decades he has been studying the application of image processing techniques to copyright protection and authentication of multimedia, and the possibility of processing signals that have been previously encrypted without decrypting them. Lately he has been working on theoretical and practical aspects of adversarial signal processing with a particular focus on adversarial multimedia forensics. He is author/co-author of about 350 papers published in international journals and conference proceedings, and holds five patents in the field of digital watermarking and image authentication. He is co-author of the book “Watermarking Systems Engineering: Enabling Digital Assets Security and other Applications”, published by Dekker Inc. in February 2004. He participated to several National and International research projects on diverse topics, including computer vision, multimedia signal processing, remote sensing, digital watermarking, multimedia forensics.
			He has been the Editor in Chief of the IEEE Transactions on Information Forensics and Security for the years 2015-2017. He was the funding editor of the EURASIP Journal on Information Security. He has been serving as associate editor of many journals including several IEEE Transactions. Prof. Barni has been the chairman of the IEEE Information Forensic and Security Technical Committee (IFS-TC) from 2010 to 2011. He was the technical program chair of ICASSP 2014. He was appointed DL of the IEEE SPS for the years 2013-2014. He is the recipient of the Individual Technical Achievement Award of EURASIP for 2016. He is a fellow member of the IEEE, AAIA, and EURASIP.
		\end{IEEEbiography}
		%
		%
		%
		%
		
	\end{document}